\documentclass[sigconf]{acmart}
\usepackage{amsmath}
\usepackage[table]{xcolor}
\usepackage{multirow}
\usepackage{bm}
\usepackage{makecell}
\usepackage{pifont}
\definecolor{DarkRed}{RGB}{178,34,34} % 手动定义暗红色

\usepackage{microtype}
\usepackage{graphicx}
\usepackage{subfig}
\usepackage{booktabs} % for professional tables
\usepackage{float} 
\usepackage{subcaption}

\usepackage{caption}
\usepackage{mathtools}
\usepackage{amsfonts}
\usepackage{dsfont}
\usepackage{url}
\usepackage{xcolor}
\usepackage{url}
\usepackage{amsmath,amsthm}
\usepackage{booktabs}
\usepackage{graphicx}
\usepackage{wrapfig}
\usepackage{algorithmic}
\usepackage{algorithm}
\usepackage{gensymb}

\newcommand{\ie}{\textit{i.e.}}
\newcommand{\eg}{\textit{e.g.}}

\AtBeginDocument{%
  }

\setcopyright{acmlicensed}
\copyrightyear{2018}
\acmYear{2018}
\acmDOI{XXXXXXX.XXXXXXX}
\acmConference[Conference acronym 'XX]{}{June 03--05,
  2018}{Woodstock, NY}
\begin{document}

% \title{TimeRadar: Time–Frequency Manipulation for Generalist Time Series Anomaly Detection}

\title{TimeRadar: A Domain-Rotatable Foundation Model for Time Series Anomaly Detection}
% \title{TimeRadar: Foundation Models for Time Series Anomaly Detection with Rotatable Deviation Learning}

\author{Hui He}
\authornote{Both authors contributed equally to this research.}
\email{huihe@smu.edu.sg}

\affiliation{%
 \institution{School of Computing and Information
Systems\\ Singapore Management University}
  \country{Singapore}
}
\author{Hezhe Qiao}
\authornotemark[1]
\email{hezheqiao.2022@phdcs.smu.edu.sg}
\affiliation{%
 \institution{School of Computing and Information
Systems\\ Singapore Management University}
  \country{Singapore}
}

 \author{Yutong Chen}
 \email{chenyutong@cigit.ac.cn}
\affiliation{%
 \institution{Chongqing Institute of Green and Intelligent Technology, University of Chinese Academy of Sciences}
   \country{Chongqing, China}
}

 \author{Kun Yi}
 \email{kunyi.cn@gmail.com}
\affiliation{%
 \institution{State Information Center}
   \country{Beijing, China}
}

 \author{Guansong Pang}
 \email{gspang@smu.edu.sg}
\authornote{Corresponding author: Guansong Pang}
\affiliation{%
 \institution{School of Computing and Information
Systems\\ Singapore Management University}
   \country{Singapore}
}

\renewcommand{\shortauthors}{Trovato et al.}

\begin{abstract}
% Foundation models for time series anomaly detection (TSAD)
% , pre-trained on large-scale multi-domain datasets and directly transferable to downstream scenarios, 
% have attracted growing attention due to their broad applicability in areas such as industrial monitoring and cybersecurity.
% Existing time series foundation models—including general-purpose and task-specific ones—are typically built in the purely time domain or frequency domain, where certain anomalies are difficult to distinguish due to the absence of unified time–frequency characterization.
Current time series foundation models (TSFMs) primarily focus on learning prevalent and regular patterns within a predefined time or frequency domain to enable supervised downstream tasks (\eg, forecasting). Consequently, they are often ineffective for inherently unsupervised downstream tasks—such as time series anomaly detection (TSAD), which aims to identify rare, irregular patterns.
% --both general-purpose and task-specific--are typically trained in a single domain (pure time or pure frequency) and often rely on self-supervised objectives such as reconstruction. 
% In these domains, 
This limitation arises because such abnormal patterns can closely resemble the regular patterns when presented in the same time/frequency domain.
% normal variations, and cross-dataset/domain conflicts are not explicitly handled, which limits the ability to learn truly generalist and transferable representations across diverse datasets.
% leading to overlapping representations
% since normal series often exhibit large intra-class variations (e.g., trend/seasonality changes and phase shift s), making anomalous deviations shape-wise similar to normal fluctuations and may be reconstructed almost as well as normal ones, leading to overlapping representations. 
To address this issue,
% these limitations, in this paper, we empirically reveal that anomalies can manifest more distinctive, highly discriminative patterns in the continuous fractional time–frequency domain, providing stronger representational power for separating them from normal behavior. To leverage the discriminative expressiveness of the unified time–frequency domain, 
we introduce \textbf{TimeRadar}, an innovative TSFM built in a fractional time–frequency domain to support generalist TSAD across diverse unseen datasets. 
% One key insight is that anomalies from different datasets show discriminative patterns in the different fractional order of the continuous time–frequency domain.
Our key insight is that rotating a time series into a data-dependent fractional time–frequency representation can adaptively differentiate the normal and abnormal signals across different datasets.
% angle effectively highlights anomalies that are subtle or indistinguishable in the pure time or frequency domain, making them more salient and easier to detect.
To this end, a novel component, namely Fractionally modulated Time-Frequency Reconstruction (\textbf{FTFRecon}), is proposed in TimeRadar to leverage a learnable fractional order to rotate the time series to the most pronounced angle between a continuous time and frequency domain for accurate data reconstruction.
% so that normal data can be well reconstructed
% It aligns each dataset to a 
This provides adaptive data reconstruction in an optimal time–frequency domain for each data input, enabling effective differentiation of the unbounded abnormal patterns from the regular ones across datasets, including unseen datasets. 
% This reduces domain mismatch and makes the pre-trained knowledge more transferable, improving generalization to previously unseen datasets. By reconstructing time series in a rotatable time–frequency domain, the model can learn more generalizable normal/abnormal patterns from the resulting time–frequency representations.
To allow TimeRadar to model local abnormality that is not captured by the global data reconstruction,
% , yielding a continuous time–frequency plane. 
% Reconstructing these time–frequency features encourages the model to learn a more discriminating representation. To further enlarge the separation between normal and abnormal patterns in representation space,
we further introduce a Contextual Deviation Learning (\textbf{CDL}) component to model the local deviation of the input relative to its contextual time series data in the rotatable domain.
% normal clusters while pushing anomalies away with a clear margin. 
Extensive experiments on eight popular TSAD benchmarks
% and the UCR benchmark 
demonstrate that TimeRadar consistently outperforms a variety of conventional and TSFM-based competing methods, delivering average gains of 10.5\% and 29.4\% in AUC-R and AUC-P, respectively. Our code will be available at \url{https://github.com/mala-lab/TimeRadar}.

\end{abstract}

%    <concept>
%        <concept_id>10010147.10010257</concept_id>
%        <concept_desc>Computing methodologies~Machine learning</concept_desc>
%        <concept_significance>500</concept_significance>
%        </concept>
%    <concept>
%        <concept_id>10002951.10003227.10003351</concept_id>
%        <concept_desc>Information systems~Data mining</concept_desc>
%        <concept_significance>500</concept_significance>
%        </concept>
%  </ccs2012>

% \ccsdesc[500]{Computing methodologies~Machine learning}
% \ccsdesc[500]{Information systems~Data mining}
\keywords{Time Series, Foundation Model, Anomaly Detection}



\maketitle

\section{Introduction}
Time series anomaly detection (TSAD), which seeks to identify rare, irregular patterns, is crucial for a wide range of applications, including infrastructure security, intelligent operation and maintenance, and space exploration \cite{caiself, yang2023dcdetector, DBLP:conf/icml/LiCCWTZ23,zhou2023detecting, liang2024foundation, chen2024self, DBLP:conf/icml/ZhouP0HG0XPL25}. Despite recent progress driven by deep learning methods, existing methods typically train a separate model for each dataset, resulting in limited generalization across target datasets and degraded anomaly detection performance in scenarios with scarce training data. Motivated by the success of \textit{foundation models} (FMs) in natural language processing~\cite{brown2020language}, 
some recent studies \cite{DBLP:conf/icml/DasKSZ24,ansari2024chronos,DBLP:conf/icml/WooLKXSS24,DBLP:conf/icml/0014LWALZL0SXS25,DBLP:conf/iclr/ShiWNLYWJ25,he2025sempo, DBLP:conf/icml/FawSZD25, meyer2025time} have established similar paradigms for 
% general-purpose
time series FMs (TSFMs).
% that support multiple tasks, \eg, classification~\cite{zhou2023one, gao2024units} and forecasting~\cite{DBLP:conf/nips/EkambaramJ0MNGR24, DBLP:conf/icml/DasKSZ24, he2025sempo}. 

\begin{figure}
 \centering
 \includegraphics[width=0.48\textwidth]{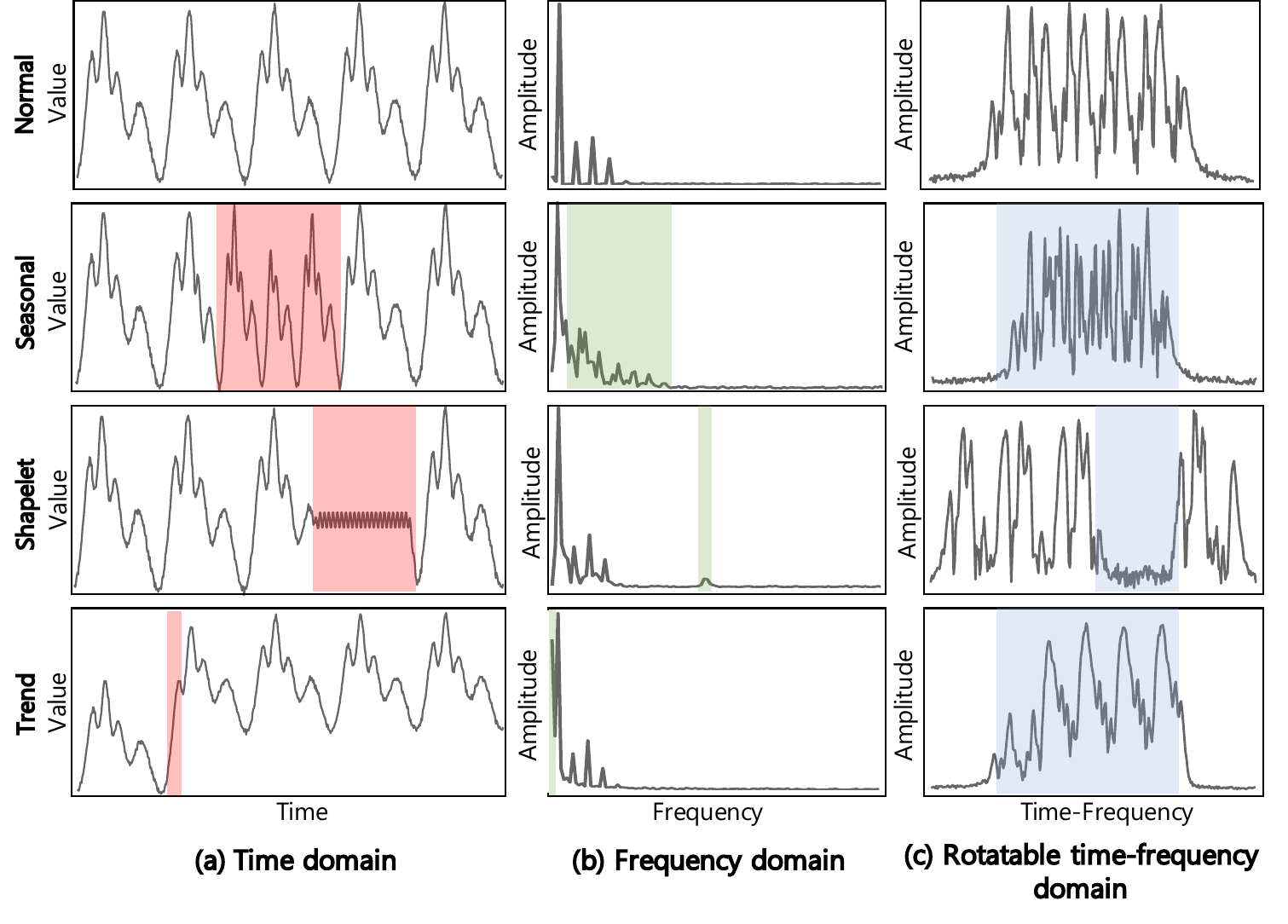}
 \caption{Normal patterns vs. three types of anomaly patterns (trend, shapelet, and seasonal anomalies, highlighted by the colored shadows) across three domains on synthetic data. (a) In the \textit{time domain}, trend and seasonal anomalies exhibit changes that are highly similar to normal variations,  whereas shapelet anomalies are more identifiable. (b) In the \textit{frequency domain}, shapelet anomalies exhibit changes that are highly similar to normal variations whereas trend and seasonal anomalies produce clearer deviations from the normal patterns. (c) In our proposed \textit{rotatable time–frequency domain}, all three anomaly types exhibit pronounced deviations from the normal time series.
}
\label{fig:example}
\vspace{-1em}
\end{figure}

Despite the substantial progress,
% on TSFMs, 
% works have explored anomaly-specific FMs tailored to TSAD \cite{DBLP:conf/iclr/ShentuL0SRPYG25, lan2025towards}. Meanwhile, 
existing TSFMs primarily focus on learning prevalent and regular patterns 
% within a predefined time or frequency domain 
to enable supervised downstream tasks \cite{DBLP:conf/icml/DasKSZ24,ansari2024chronos,DBLP:conf/icml/WooLKXSS24,DBLP:conf/icml/0014LWALZL0SXS25,DBLP:conf/iclr/ShiWNLYWJ25,he2025sempo,DBLP:conf/icml/0004Q000RP0G25,gao2024units,DBLP:conf/nips/EkambaramJ0MNGR24, hua2026diversified}. However, this objective is mismatched to those in inherently
unsupervised downstream tasks, especially the TSAD task where we are interested in detecting rare abnormal events whose patterns are unbounded and can differ significantly from each other across datasets, \eg, the three different types of anomalies in Fig. \ref{fig:example}. Furthermore, these TSFMs are often trained in a predefined time or frequency domain, making them more difficult in detecting these diverse anomalies. 
% Among existing efforts, DADA~\cite{DBLP:conf/iclr/ShentuL0SRPYG25} is a representative generalist TSAD model, pre-trained on large-scale multi-domain datasets using a self-supervised objective and subsequently transferable to a wide range of downstream scenarios. Although DADA has made notable progress toward TSAD FMs, 
This is because certain abnormal patterns can closely mimic normal variations when presented in the same time/frequency
domain.
% In the time domain, in particular, normal behavior may be widely dispersed, making some anomalies weak, subtle, and difficult to distinguish.
For example, as shown in Fig. \ref{fig:example} (a), while shapelet anomalies differ markedly from normal time series, trend and seasonal anomalies induce changes that closely resemble normal temporal variations in the time domain. 
% Moreover, they do not explicitly address cross-dataset/domain conflicts
% and domain mismatch, making it difficult to learn truly generalist,
% transferable representations across diverse datasets.
As a result, time-domain FMs often struggle to localize all these abnormal patterns \cite{DBLP:conf/icml/DasKSZ24, DBLP:conf/nips/EkambaramJ0MNGR24, DBLP:conf/icml/WooLKXSS24, ansari2024chronos, DBLP:conf/iclr/ShiWNLYWJ25, DBLP:conf/iclr/ShentuL0SRPYG25, DBLP:conf/icml/0014LWALZL0SXS25, DBLP:conf/icml/FawSZD25, lan2025towards}.
% , leading to overlapping representations between normal and abnormal samples.
A similar issue exists in frequency-domain FMs \cite{he2025sempo,DBLP:conf/icml/0004Q000RP0G25,DBLP:conf/ijcai/Wang00B0P25},
% To mitigate this issue, several works shift the focus to the frequency domain, for example, by applying spectral transforms to highlight characteristic frequency components \cite{DBLP:conf/iclr/WuQL0HGXY25, he2025sempo, fang2024temporal}. However, purely
since frequency domain modeling can weaken the temporal patterns. This is illustrated in Fig.~\ref{fig:example} (b), where the shapelet and trend anomalies exhibit only minor differences in the global spectrum. 
% Consequently, the representations in the frequency domain are often less effective.

Very recently there have been some efforts on building FMs specifically for TSAD, such as DADA~\cite{DBLP:conf/iclr/ShentuL0SRPYG25} that is pre-trained to detect anomalies from large-scale multi-domain datasets. However, it is trained in the time domain, suffering from the similar issues as the other time-domain TSFMs.

To address these limitations, we introduce \textbf{TimeRadar}, an innovative FM built in a rotatable time–frequency domain for enabling generalist TSAD across unseen datasets. The key insight of TimeRadar is that rotating a time series into a data-dependent fractional time–frequency representation can adaptively differentiate the normal and abnormal signals across different datasets.
% leveraging a learnable fractional order to rotate time series into a data-dependent time–frequency representation can adaptively differentiate the normal and abnormal patterns across different datasets, rendering anomalies more salient and easier to detect.
As illustrated in Fig. \ref{fig:example} (c), all three anomaly types exhibit clear and pronounced deviations from the normal time series in a rotatable time-frequency domain (see App.~\ref{app:anomaly_visualization} for more visualizations of anomalies under different rotation angles). Motivated by this insight, we propose \textbf{FTFRecon}, a \underline{F}ractionally modulated \underline{T}ime-\underline{F}requency \underline{Recon}struction component in TimeRadar, in which a learnable fractional order is optimized to rotate time series to the most pronounced angle between the continuous time and frequency domain for minimizing data reconstruction errors. 
% In the rotated space, FTFRecon performs adaptive masked reconstruction through a fractionally modulated encoder 
In doing so, FTFRecon can adaptively reconstruct each data input
% operates on time–frequency features and with patching and complementary masking, and is optimized using an  
in an optimal time–frequency domain,
% for each input. This design 
enabling effective differentiation of the unbounded abnormal patterns from the regular ones across datasets.
% , as well as those previously unseen during pretraining.
% improving generalization to previously unseen datasets. 
% It aligns each dataset to a dataset-specific orientation, reducing domain mismatch and making the pre-trained knowledge more transferable, improving generalization to previously unseen datasets. 

To capture local abnormalities that global data reconstruction fails to characterize adequately, we further introduce a Contextual Deviation Learning (\textbf{CDL}) component in the rotatable domain, which models the patch-wise deviation of each input with respect to its contextual time series data, promoting compactness among normal samples while enlarging their separation from anomalies with a clear margin. In summary, our contributions are as follows:
\begin{itemize}
\item We introduce the TSAD-oriented FM, \textbf{TimeRadar}. To the best of our knowledge, TimeRadar is the first TSFM that learns to adaptively detect diverse anomalies across different datasets in a domain-rotatable time–frequency space. 

\item TimeRadar is instantiated by two novel components, including FTFRecon and CDL. 
% propose a Fractionally Modulated Time–Frequency Reconstruction component that automatically
\textbf{FTFRecon} is optimized to adaptively identify and rotate the time series to the most pronounced rotation angle in the continuous time–frequency domain, so that the data reconstruction can be performed in the optimal fractional domain for each data input. This helps learn more discriminative TSAD patterns.
% \item The proposed Contextual Deviation Learning 
\textbf{CDL} complements the data reconstruction objective in FTFRecon to model local abnormalities by explicitly enforcing large separation between normal and abnormal classes based on the deviation of each input from its contextual time-series.

\item Extensive experiments on eight popular TSAD benchmarks demonstrate that TimeRadar delivers the best overall performance, consistently and substantially outperforming both TSFM-based methods and conventional deep TSAD baselines under both zero-shot and few-shot inference settings.
% , with average gains of 29.40\% in AUC-P and 10.50\% in AUC-R.

\end{itemize}

\section{Related Work}

\subsection{Time Series Foundation Models}
TSFMs are typically pre-trained on a diverse collection of datasets across multiple domains and can be directly adapted to unseen target domains via zero-shot inference or few-shot fine-tuning \cite{DBLP:conf/iclr/NieNSK23,he2025sempo}. 
Existing TSFMs can be roughly categorized into those operating purely in the time or frequency domain and those built in the joint time–frequency domain.

\noindent \textbf{TSFMs in the time/frequency domain.}
TSFMs have been extensively studied under time-domain modeling
% , where they are typically trained with self-supervised objectives (e.g., reconstruction or forecasting) 
\cite{ansari2024chronos, DBLP:conf/iclr/ShiWNLYWJ25, DBLP:conf/iclr/ShentuL0SRPYG25, DBLP:conf/icml/0014LWALZL0SXS25, DBLP:conf/icml/FawSZD25, lan2025towards}.
The time-domain-based TSFMs can be categorized into encoder-only architectures with masked pre-training objectives \cite{goswami2024moment, DBLP:conf/icml/WooLKXSS24}, decoder-only architectures \cite{rasul2023lag, DBLP:conf/icml/DasKSZ24, DBLP:conf/iclr/ShiWNLYWJ25}, and encoder-decoder architectures \cite{liang2024foundation, ansari2024chronos, ansari2025chronos2}.
% Encoder-only architectures focus on leveraging self-supervised learning to capture intricate temporal dependencies. Decoder-only architectures are typically employed under a GPT-style causal modeling paradigm for autoregressive forecasting \cite{rasul2023lag, DBLP:conf/icml/DasKSZ24, DBLP:conf/iclr/ShiWNLYWJ25}. Encoder-decoder architecture that tokenizes numerical time series via scaling and quantization into a fixed vocabulary \cite{ansari2024chronos, ansari2025chronos2}.
% These general FMs primarily focus on learning prevalent and regular patterns within a predefined time or frequency domain. Therefore, they are often ineffective for inherently unsupervised downstream TSAD, which aims to identify rare, irregular patterns.
To capture periodicity changes and energy redistribution patterns, another line of work introduces spectral representations and constructs TSFMs in the frequency domain  
\cite{DBLP:conf/iclr/WuQL0HGXY25, chen2024learning, zhang2025frect}. Spectral transforms, \eg, Fourier Transform~\cite{DBLP:conf/nips/0001FZHHL024,DBLP:conf/kdd/000100CWHL0W025}, can emphasize characteristic frequency components, making certain anomalies easier to detect. 
However, in the single time domain or frequency domain, abnormal patterns can closely resemble the regular patterns, often making the TSFMs ineffective for TSAD.
% inherently unsupervised downstream TSAD, which aims to identify rare, irregular patterns.

% where abnormal patterns can closely mimic normal variations, leading to non-discriminative representations between normal and abnormal samples during reconstruction. In contrast, TimeRadar rotates each time series to the most pronounced angle between a continuous time and frequency domain, enabling lossless reconstruction. This encourages the model to learn more discriminative and transferable normal/abnormal patterns during reconstruction.

\noindent \textbf{TSFMs in the time-frequency domain.}
There are relatively few studies that investigate TSFMs from a time-frequency perspective \cite{fang2024temporal, DBLP:conf/iclr/WuQL0HGXY25, DBLP:conf/icde/SunPY0HY24}. They typically fuse time-domain and frequency-domain features into a time–frequency representation and capture how signal energy is distributed across time and frequency.
However, these methods rely on a straightforward fusion of time- and frequency-domain features instead of learning a continuous
time–frequency representation. In contrast, TimeRadar builds FMs by
rotating time series to an input-dependent angle—to better align datasets and capture transferable normal/abnormal patterns to unseen datasets.

% Although some prior studies have investigated time–frequency representations for forecasting or conventional TSAD tasks \cite{zhang2022tfad, nam2024breaking, DBLP:conf/ijcai/Wang00B0P25}, most of them rely on a straightforward fusion of time- and frequency-domain features. In contrast, building FMs for TSAD calls for learning a unified, continuous time–frequency representation—e.g., one that can be adaptively rotated to an input-dependent angle—to better align domains and capture transferable normal/abnormal patterns to unseen datasets.

\subsection{Time Series Anomaly Detection}
Most conventional TSAD methods rely on handcrafted statistics or density-based criteria, including one-class classification \cite{xu2024calibrated} and neighborhood-based approaches \cite{nakamura2020merlin}.  Recently deep TSAD methods receive much attention, especially in the time domain, which include representation learning based methods \cite{campos2021unsupervised, zhao2022comparative, kim2023contrastive,yang2023dcdetector} and prediction methods  \cite{su2019robust, pang2021deep}.
Representation learning methods aim to reconstruct the time series and leverage the reconstruction error as the anomaly score, or leverage the contrastive learning to learn discriminating representations for normal and abnormal time series \cite{kim2023contrastive,yang2023dcdetector, darban2026genias}. Prediction-based methods detect anomalies by learning a forecasting model from historical observations, where anomaly scores are typically computed using the forecasting residual between the predicted value and the observation \cite{DBLP:conf/ijcai/LiuLLTZ25, shen2024afmf}. Although these deep learning-based models have achieved promising results, they are trained and evaluated on the same dataset, which limits their ability to generalize robustly across diverse TSAD datasets. Recently there have been some studies on TSAD-oriented FMs~\cite{DBLP:conf/iclr/NieNSK23,lan2025towards,DBLP:conf/iclr/ShentuL0SRPYG25}, but they operate in the time domain only, suffering from less discriminative  representations on different datasets. In contrast, TimeRadar addresses this issue with a domain-rotatable time-frequency space.
% develops an FM for generalist TSAD by rotating time series to the most pronounced angle between a continuous time and frequency domain, enabling effective differentiation of the rare, unbounded abnormal patterns from the regular ones across datasets, including unseen datasets. 

% Some recent studies have also proposed anomaly-specific time series anomaly detection FMs~\cite{DBLP:conf/iclr/NieNSK23,lan2025towards}. However, these methods primarily operate in the pure time domain, where anomalies are identified by deviations in temporal patterns.
% In the pure time domain, the abnormal patterns can closely mimic normal variations, leading to non-discriminative representations between normal and abnormal samples.

% rotates each time series to the most pronounced angle between a continuous time and frequency domain, enabling lossless reconstruction. This encourages the model to learn more discriminative and transferable normal/abnormal patterns during reconstruction.

\section{The Proposed TimeRadar}

\subsection{Problem Statement}
\noindent \textbf{Time series anomaly detection (TSAD).}
Given a multivariate time series input $\mathbf{X}=[X_1,X_2,...,X_T] \in \mathbb{R}^{T \times C}$ with the number of variates $C$ and the observation length $T$, the task of TSAD is to output $\hat{\mathbf{Y}}= [y_1,y_2,...,y_T] \in \mathbb{R}^{T \times 1}$, where $y_t \in \{0,1\}$ denotes whether the observation $X_t \in \mathbb{R}^C$ of the $t$-th time step is anomalous.

\noindent \textbf{Foundation model for Generalist TSAD.}
This work focuses on developing a foundation model (FM) for enabling generalist TSAD. The model is pre-trained on a collection of $I$ time series datasets from diverse domains, $\mathcal{T}_{\mathrm{train}}=\{\mathcal{D}_{\mathrm{train}}^1,\mathcal{D}_{\mathrm{train}}^2,...,\mathcal{D}_{\mathrm{train}}^I\}$, where $\mathcal{D}_{\mathrm{train}}^i=\{\mathbf{X}^i \in \mathbb{R}^{T^{(i)} \times C^{(i)}},\mathbf{Y}^i \in \mathbb{R}^{T^{(i)} \times 1}\}$ represents the $i$-th dataset, $C^{(i)}$ and $ \mathbb{R}^{T^{(i)}}$ are the number of variates and time points. For downstream evaluation, we consider two settings, including zero- and few-shot anomaly detection. In the few-shot setting, the pre-trained model is further adapted using a small target dataset $\mathcal{D}_{\mathrm{tune}}$, where $|\mathcal{D}_{\mathrm{tune}}^\prime| \ll |\mathcal{T}_{\mathrm{train}}|$, and evaluated on an unseen $\mathcal{D}_{\mathrm{test}}^\prime$ from the same domain with $\mathcal{D}_{\mathrm{tune}}^\prime$. In the zero-shot setting, the model is tested directly on $\mathcal{D}_{\mathrm{test}}^\prime$ without any additional adaptation. 
% In both cases, $\mathcal{T}_{\mathrm{train}}$, $\mathcal{D}_{\mathrm{tune}}$ and $\mathcal{D}_{\mathrm{test}}$ are pairwise disjoint, i.e., $\mathcal{T}_{\mathrm{train}} \cap \mathcal{D}_{\mathrm{tune}} \cap \mathcal{D}_{\mathrm{test}}=\emptyset$. 
Following standard zero- and few-shot settings, $\mathcal{D}_{\mathrm{tune}}^\prime$ and $\mathcal{D}_{\mathrm{test}}^\prime$ are drawn from entirely new domains that are not presented in $\mathcal{T}_{\mathrm{train}}$, \ie, $ \mathcal{D}_{\mathrm{tune}}^\prime,\mathcal{D}_{\mathrm{test}}^\prime \not\subseteq \mathcal{T}_{\mathrm{train}}$. Since datasets from diverse domains typically have varying numbers of variates $C^{(i)}$, we apply channel independence~\cite{DBLP:conf/iclr/NieNSK23,DBLP:conf/iclr/ShentuL0SRPYG25} to decompose multivariate inputs into univariate series, thus enabling any-variate anomaly detection.

\subsection{Overview of the Proposed TimeRadar}
\begin{figure*}
\centering
% Requires \usepackage{graphi
\includegraphics[width=0.95\textwidth]{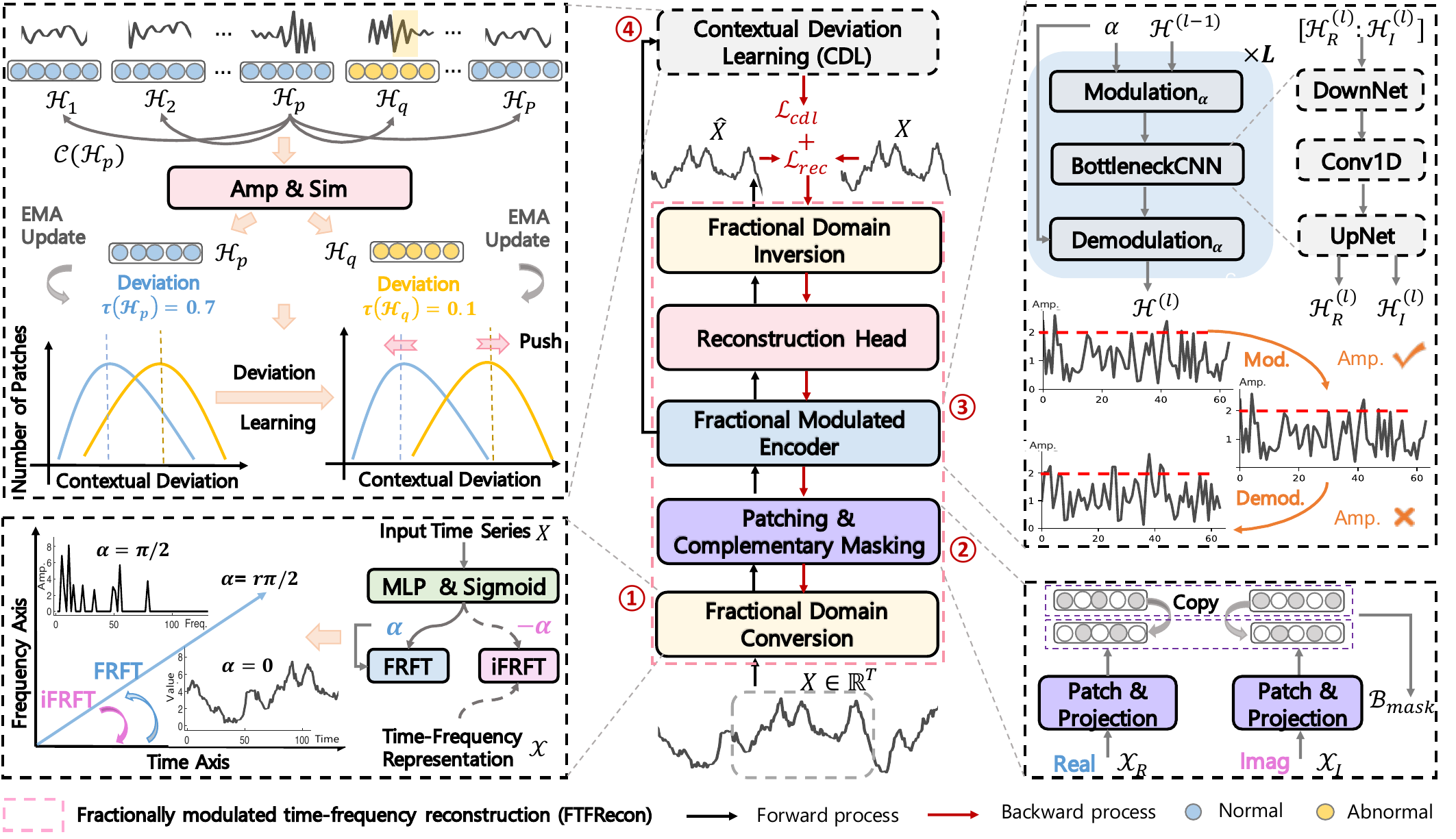}\\
\caption{Overview of TimeRadar. Given a time series $X$, its FTFRecon module firstly \textcolor{DarkRed}{\ding{172}} learns an input-dependent adaptive rotation angle $\alpha$ and accordingly transforms $X$ into a flexible and expressive fractional time–frequency domain. After \textcolor{DarkRed}{\ding{173}} patching and complementary masking, the complex-valued time series patches are fed into the \textcolor{DarkRed}{\ding{174}} fractionally modulated encoder and a reconstruction head. To complement the data reconstruction with local abnormality modeling, the \textcolor{DarkRed}{\ding{175}} CDL module models the deviation of each patch relative to its contextual patches in the time–frequency hidden space via a margin learning loss. During pre-training, TimeRadar is jointly optimized with the reconstruction loss $\mathcal{L}_{rec}$ and the margin loss $\mathcal{L}_{cdl}$.
}
\label{fig:framework}
\end{figure*}

As illustrated in Fig. \ref{fig:framework}, TimeRadar is built upon an encoder-only architecture, which comprises two key components: fractionally modulated time-frequency reconstruction (FTFRecon) and contextual deviation learning (CDL). FTFRecon consists of four steps: fractional domain conversion/inversion, patching $\&$ complementary masking, a fractionally modulated encoder, and a reconstruction head. During pre-training, the model is optimized using both a reconstruction loss and a margin-based deviation loss, as detailed in Sec. \ref{pretraining}.
% TimeRadar first learns an input-dependent rotation angle that transforms each time series input into a continuous time–frequency domain. Then it reconstructs the time-frequence representation under patching and complementary masking.
% To make the model more adaptive to the data, TimeRadar further leverages CDL to explicitly enlarge the separation between normal and abnormal patterns in the learned representation space.   
% During inference, TimeRadar uses the variance across reconstructed values at the same time point as the anomaly score, since normal samples can be reconstructed stably with low variability, whereas abnormal samples tend to yield more unstable reconstructions with higher variance.
Below we introduce TimeRadar in detail.

% \subsection{Time–frequency Learning}
% FRFT generalizes the traditional Fourier Transform (FT) to the entire spatial-frequency domain with an additional parameter known as the order $r$, which is related to a signal rotation angle in the spatial-frequency plane.

\subsection{FTFRecon: Fractionally Modulated Time-Frequency Reconstruction}
FTFRecon is designed to reconstruct each data input in an optimal fractional time–frequency domain, enabling the learning of normal patterns of the data inputs in a domain-adaptive manner.

\noindent \textbf{Fractional domain conversion/inversion.}
The Fractional Fourier Transform (FRFT)~\cite{DBLP:conf/nips/YuHLZZ23} generalizes the traditional Fourier Transform (FT)~\cite{DBLP:conf/nips/YiZFWWHALCN23} by enabling a continuous $r$-th order counterclockwise rotation in the entire time-frequency plane.
This property is particularly advantageous for time series analysis, as FRFT provides a family of intermediate representations that bridge the time and frequency domains. Such representations can well align the dominant time–frequency trajectory of normal signals, while abnormal signals remain misaligned and thus exhibit more dispersed energy distributions, see Fig. \ref{fig:example}, which facilitates the discrimination between normal and abnormal patterns in anomaly detection tasks. 

Given a univariate time series $X \in \mathbb{R}^T$, we first apply instance normalization~\cite{DBLP:conf/iclr/KimKTPCC22} and the resulting normalized sequence is then transformed along the temporal dimension using the FRFT to obtain $\mathcal{X}(v_r)$, formulated as:
% after instance normalization~\cite{}, the normalized input is converted into the fractional time-frequency domain by applying
\begin{equation} \label{eqn1}
    \mathcal{X}(v_r)=\int_{-\infty}^{\infty} K_r(v_0, v_r)X(v_0)\mathrm{d}v_0,
\end{equation}
where $v_r$ denotes the fractional time-frequency variable at the $r$-th fractional order, $v_0$ is the integral variable in the time domain, and the kernel function $K_r(v_0, v_r)$ is defined as: 
\begin{equation} \label{eqn2}
    K_r(v_0, v_r)=\begin{cases}
        A_\alpha e^{j\Phi_\alpha(v_0, v_r)}, & \alpha \neq n\pi \\
        \delta(v_r - v_0), & \alpha = 2n\pi \\
        \delta(v_r + v_0), & \alpha = (2n+1)\pi 
        
    \end{cases}
\end{equation}
where $j$ is the imaginary unit. The FRFT kernel is characterized by the amplitude term $A_\alpha=\sqrt{\frac{1-j\cot\alpha}{2\pi}}$ and the quadratic phase function $\Phi_\alpha(v_0, v_r)=v_0^2\cot\alpha/2-v_rv_0\csc\alpha + v_r^2\cot\alpha/2$.
% with $\alpha=\frac{r\pi}{2}$. 
$\delta(\cdot)$ denotes the Dirac delta function,
% \cite{dirac1927physical}
indicating domain identity or domain reversal in the transformation. To capture domain-specific time-frequency characteristics across multi-domain data, we further parameterize the fractional order $r$ as an input-dependent variable learned from the input series $X$, which is then mapped to the rotation angle $\alpha$, thereby enabling adaptive time-frequency rotation, defined as: 
% \begin{equation} \label{eqn3}
%     r=\mathrm{Sigmoid}(\mathrm{MLP}(\frac{1}{C}\sum_{c=1}^C\mathbf{X}_{:,c}))
% \end{equation}
\begin{equation} \label{eqn3}
    \alpha = \frac{r\pi}{2}, \ r=\mathrm{Sigmoid}(\mathrm{MLP}(X)).
\end{equation}

After the reconstruction of the masked series in the fractional time–frequency domain, we convert $\mathcal{X}$ back into the time domain using the following inverse conversion formulation:
\begin{equation} \label{eqn4}
    X(v_0)=\int_{-\infty}^{\infty} K_{-r}(v_0, v_r)\mathcal{X}(v_r)\mathrm{d}v_r.
\end{equation}

\noindent \textbf{Patching and complementary masking.}
Considering that significant amplitude disparities may exist among different components of the input data, the model can be dominated by high-magnitude components, while low-magnitude components are easily suppressed~\cite{DBLP:conf/kdd/Piao0MMS24,DBLP:conf/nips/0001FZHHL024,DBLP:conf/iclr/WuQL0HGXY25}. However, low-magnitude components often capture fine-grained time–frequency deviations that are highly sensitive to abnormal dynamics. 
Therefore, 
% inspired by the strategy of frequency refinement~\cite{DBLP:conf/kdd/Piao0MMS24},
we introduce a patching and projection step in the fractional domain to extract localized time-frequency patterns. 
Concretely, given the fractional domain conversion output $\mathcal{X} \in \mathbb{C}^{T}$, we divide it into $P$ non-overlapping patches of equal length $T_p$ with stride $s$, where $P=[T-T_p]/s+1$. Each patch is then projected into a high-dimensional space of dimension $D_p$ through the projection layer, which is formalized as:
\begin{equation} \label{eqn5}
    \mathcal{B}_R=\mathrm{Projection}(\mathrm{Patch}(\mathcal{X}_R)), \mathcal{B}_I=\mathrm{Projection}(\mathrm{Patch}(\mathcal{X}_I)),
\end{equation}
where $\text{Projection}(\cdot)$ is implemented by a single linear layer, with $\mathcal{X}_R$ and $\mathcal{X}_I$ denoting the real and imaginary parts of $\mathcal{X}$, respectively.

% Considering that masking-based modeling~\cite{DBLP:conf/icml/DasKSZ24,he2025sempo,DBLP:conf/iclr/ShentuL0SRPYG25} facilitates the leaning of normal temporal dependencies by 
% % deliberately occluding a subset of patches and then
% reconstructing the missing information from the remaining context, 
To enforce the model see global contextual dependencies, we adopt random patch masking~\cite{DBLP:conf/icml/DasKSZ24,he2025sempo,DBLP:conf/iclr/ShentuL0SRPYG25} in the fractional time-frequency domain. 
To preserve phase consistency between the real parts and imaginary parts, this masking is implemented as a shared complementary strategy, where the same patch-wise binary mask and its complement are consistently applied to both parts, yielding dual complementary views of the time-frequency representation.  
% randomly masking a portion of patches along the temporal dimension, formalized by:
Concretely, $\mathcal{B}_R$ and $\mathcal{B}_I$ can be combined to from the complex-valued representation $\mathcal{B}=\mathcal{B}_R+j\mathcal{B}_I \in \mathbb{C}^{P \times D_p}$. Given the $w$-th binary mask $\mathcal{M}_w \in \{0,1\}^P$, two complementary masked representations are then constructed as $\mathcal{M}_w \odot \mathcal{B}$ and $(1-\mathcal{M}_w) \odot \mathcal{B}$, where $\odot$ denotes element-wise multiplication. The final masked output is obtained by concatenating $W$ complementary mask pairs:

\begin{equation} \label{eqn7}
    \mathcal{B}_{\mathrm{mask}}=\mathrm{Concat}[\mathcal{M}_w \odot \mathcal{B}, (1-\mathcal{M}_w) \odot \mathcal{B}]_{w=1}^W.
\end{equation}

% concatenate the real and imaginary parts and perform a fractional-domain convolutional learning process.

\noindent \textbf{Fractionally modulated encoder.}
An fractionally modulated encoder is designed to model complex-valued time-frequency representations by performing convolution under rotated time-frequency coordinates. Since local phase misalignment may still exist across patches after rotation, modulation is introduced to reduce such inconsistencies and make the representations more compatible with convolutional weight sharing. Given the masked complex-valued representation $\mathcal{B}_{mask} \in \mathbb{C}^{W \times P \times D_p}$, we first apply a stack of fractional modulation blocks, denoted as \textit{FracBlocks}, where each block consists of a modulation-convolution-demodulation procedure in the fractional time-frequency domain. Formally, the fractional modulated learning consists of $L$ \textit{FracBlocks}. For the $l$-th block, the forward propagation is defined as:
\begin{align} \label{eqn8}
    \mathcal{H}^{(l)}&=\mathrm{Modulation}_{\alpha}(\mathcal{H}^{(l-1)}), \\
    \mathcal{H}^{(l)}&=\mathrm{BottleneckCNN}(\mathcal{H}^{(l)}), \\
    \mathcal{H}^{(l)}&=\mathrm{Demodulation}_{\alpha}(\mathcal{H}^{(l)}),
\end{align}
where $\mathcal{H}^{(0)}=\mathcal{B}_{mask}$. The modulation operator applies a quadratic-phase chirp to rotate the complex-valued representation, 
% in the fractional domain, 
which is implemented by an element-wise multiplication along the $P$ dimension, \ie, $\mathrm{Modulation_{\alpha}(\mathcal{H})}=\mathcal{H} \odot_P \mathbf{e}_{\alpha}$. As shown in Fig. \ref{fig:framework}, 
% this operation alters only the phase while leaving the amplitude unchanged 
this operation rotates the phase while preserving the amplitude. Correspondingly, the demodulation operator removes the phase rotation by multiplying the complex conjugate chirp, \ie, $\mathrm{Demodulation_{\alpha}(\mathcal{H})}=\mathcal{H} \odot_P \mathbf{e}_{\alpha}^\star$, where $(\cdot)^\star$ denotes the complex conjugate. The quadratic-phase modulation function is defined as $\mathbf{e}_{\alpha}=e^{j\pi\cot\alpha\frac{(p-(p-1)/2)^2}{P}}$.

Next, the \textit{BottleneckCNN} in each \textit{FracBlock} is implemented as a real-valued dilated convolution operating on concatenated real and imaginary parts. Since concatenation doubles the representation dimensionality,
% To improve the stability and generalization of the learned complex-valued fractional modeling, 
we introduce a bottleneck into \textit{BottleneckCNN} to control parameterization, which implicitly induces a low-rank structure in the representation space. The computation of \textit{BottleneckCNN} is defined as:
\begin{align} \label{eqn9}
    \mathbf{H}^{(l)}&=\mathrm{DownNet}(\mathrm{Concat}(\mathcal{H}^{(l)}_R, \mathcal{H}^{(l)}_I)), \\
    \mathbf{H}^{(l)}&=\mathrm{Conv1D}(\mathbf{H}^{(l)}), \\
    \mathcal{H}^{(l)}_R, &\mathcal{H}^{(l)}_I =\mathrm{Chunk}(\mathrm{UpNet}(\mathbf{H}^{(l)})),
\end{align}
where $\mathrm{DownNet}(\cdot)$, $\mathrm{Conv1D}(\cdot)$, and $\mathrm{UpNet}(\cdot)$ are all implemented as 1D convolutions applied along the $P$ dimension, with $\mathrm{DownNet}: \mathbb{R}^{2*D_P} \rightarrow \mathbb{R}^{D_p}$, $\mathrm{Conv1D}(\cdot): \mathbb{R}^{D_P} \rightarrow \mathbb{R}^{D_o}$, $\mathrm{UpNet}: \mathbb{R}^{D_o} \rightarrow \mathbb{R}^{2*D_o}$. 
% the network which compresses the representation into a latent space, $\mathbb{R}^{D_P} \rightarrow \mathbb{R}^{D_i}$, where $D_i$ is the latent space dimension and , $\mathrm{UpNet}(\cdot)$ restores the representation from the latent space, $\mathbb{R}^{D_i} \rightarrow \mathbb{R}^{D_P}$. 
Then, we stack $\mathcal{H}^{(l)}_R$ and $\mathcal{H}^{(l)}_I$ to form a complex-valued $\mathcal{H} \in \mathbb{R}^{2*D_o}$ for the subsequent reconstruction head.

\noindent \textbf{Reconstruction head.}
Finally, we employ a complex-valued Multi-layer Perception (MLP)~\cite{DBLP:conf/nips/YiZFWWHALCN23,DBLP:conf/nips/0001FZHHL024} to project the modulated time-frequency representation $\mathcal{H}$ into the reconstruction space, formulated as: 
\begin{equation} \label{eqn10}
    \mathcal{X}= (\mathrm{GeLU}(\mathcal{H}\mathcal{W}_1+\mathcal{B}_1))\mathcal{W}_2+\mathcal{B}_2,
\end{equation}
where $\mathcal{W}_1$, $\mathcal{W}_2$ and $\mathcal{B}_1$, $\mathcal{B}_2$ denote the complex-valued weights and biases, and $\mathrm{GeLU}$ is the activation function. The final reconstructed signal $\hat{X} \in \mathbb{R}^{W \times T}$ is obtained by applying inverse FRFT, followed by the amplitude projection:
\begin{equation} \label{eqn11}
    \hat{X}=\mathrm{Amp}(\mathrm{iFRFT}(\mathcal{X})).
\end{equation} 
% As shown in Eq. (\ref{eqn14}), the reconstruction head generates the result $\hat{X}$ from the input series $X$, and 
The reconstruction loss is defined as:
\begin{equation}
    \mathcal{L}_{rec}=\frac{1}{T}\sum_{t=1}^T||X_t-\hat{X}_t||^2_2.
\end{equation}

\subsection{CDL: Contextual Deviation Learning}
% The representations learned by FTFRecon can effectively capture both normal and abnormal patterns, but minor noise in anomalies or contamination within normal data may still interfere with the learning process.

Local abnormalities are often overlooked by global reconstruction. To better capture such fine-grained abnormalities that global data reconstruction fails to characterize, we introduce Contextual Deviation Learning (CDL), which enlarges the separation between normal and abnormal patterns in the time–frequency representation, promoting compact normal clusters while pushing anomalies away with a clear margin. 

Given the output of the fractional modulated encoder $\mathcal{H}=\{\mathcal{H}_1, \mathcal{H}_2,...,\mathcal{H}_P\} \in \mathbb{C}^{W \times P \times D_p}$, the contextual deviation of patch $\mathcal{H}_p$, denoted as $\tau(\mathcal{H}_p)$, is defined as the aggregated differences to its context patches $\mathcal{C}(\mathcal{H}_p)$ within the same time window of length $T$:
% to capture relational discrepancies among patches:
\begin{equation} \label{eqn12}
    \tau(\mathcal{H}_p)=\frac{1}{P-1}\sum_{\mathcal{H}_q \in \mathcal{C}(\mathcal{H}_p)}\left(1-\mathrm{Sim}(\mathrm{Amp}(\mathcal{H}_p), \mathrm{Amp}(\mathcal{H}_q))\right),
\end{equation} 
where $\mathrm{Sim}(\cdot,\cdot)$ is implemented as 
% the inner product of layer-normalized amplitude.is implemented as 
the cosine similarity.
Then, we adopt a patch-wise margin loss on contextual deviation to explicitly model the contextual deviation difference for normal patches $\mathcal{H}_p^N$ and abnormal patches $\mathcal{H}_p^A$ as follows:
\begin{equation} \label{eqn13}
    \mathcal{L}_{cdl}=\max\{0, \gamma-( \tau(\mathcal{H}_p^N)-  \tau(\mathcal{H}_p^A))\},
\end{equation} 
where $\tau(\mathcal{H}_p^N)$ and $\tau(\mathcal{H}_p^A)$ denote the average contextual deviation of normal and abnormal patches, respectively:
\begin{equation} \label{eqn14}
    \tau(\mathcal{H}^N) = \frac{1}{|\mathcal{H}^N|}\sum_{\mathcal{H}_p \in \mathcal{H}^N}\tau({\mathcal{H}_p}), \;
    \tau(\mathcal{H}^A) = \frac{1}{|\mathcal{H}^A|}\sum_{\mathcal{H}_p \in \mathcal{H}^A}\tau({\mathcal{H}_p}). 
\end{equation}
As illustrated in App. \ref{app:deviation}, by applying CDL, the gap between normal and abnormal patches is larger after training than before, suggesting that CDL strengthens the contextual deviation contrast and yields a more distinct separation between normal and abnormal time series.

\subsection{Pre-Training, Fine-Tuning, and Inference} \label{pretraining}
% The pre-training loss function aims to learn the parameters of the encoder and decoder.
\noindent\textbf{Pre-Training.} During pre-training, TimeRadar is optimized under a joint objective consisting of the reconstruction loss and the context-aware margin loss. The reconstruction loss encourages the model to capture normal temporal patterns, such that anomalous samples that deviate from normal behavior yield higher reconstruction errors in the raw space. Meanwhile, the context-aware margin loss further enlarges the separation between normal and abnormal patterns in the representation space. The overall pre-training loss is formulated as:
\begin{equation} \label{eqn15}
     \mathcal{L}_{pretraining}=\mathcal{L}_{rec}+\lambda\mathcal{L}_{cdl},
\end{equation}
where $\lambda$ is a hyperparameter to control the contribution of the CDL module. During optimization, the contextual deviations $\tau(\mathcal{H}_p^N)$ and $\tau(\mathcal{H}_p^A)$ are updated using an Exponential Moving Average (EMA), defined as $\tau^{t+1}=m\tau^t+(1-m)\hat{\tau}^{(t)}$, where $m$ is the EMA momentum and $\hat{\tau}^{(t)}$ is the batch-wise estimate at training step $t$.
% The corresponding pseudocode of EMA is summarized in Algorithm 1 in Appendix.

% \subsection{Fine-Tuning and Inference}
\noindent\textbf{Fine-Tuning.} For the fine-tuning under the few-shot setting, we perform head probing~\cite{DBLP:conf/nips/EkambaramJ0MNGR24,DBLP:conf/icml/0004Q000RP0G25}, where only the reconstruction head is fine-tuned on the target dataset, with the backbone kept frozen.

\noindent\textbf{Inference.} During inference, all trainable parameters of TimeRadar are frozen. We compute the reconstruction error together with the variance of multiple reconstructed series at the same time point as the anomaly score $S(X_t)$~\cite{DBLP:conf/iclr/ShentuL0SRPYG25}, defined as:
\begin{equation} \label{eqn16}
    S(X_t) =\mathrm{Var}_W(\hat{X}_t)+||X_t-\mathrm{Mean}_W(\hat{X}_t)||_2^2.   
\end{equation}
Anomalous series are typically harder to reconstruct and thus produce more unstable, dispersed reconstructions with higher variance, whereas normal series yield consistent reconstructions with low uncertainty. For performance metrics that require binary outputs, following~\cite{DBLP:conf/iclr/ShentuL0SRPYG25,DBLP:conf/iclr/WuQL0HGXY25}, we apply SPOT~\cite{DBLP:conf/kdd/SifferFTL17} to determine the decision threshold, and a point is marked as anomalous if its anomaly score exceeds the threshold.

\section{Experiments}

\subsection{Experimental Setup}

\noindent \textbf{Datasets.}
Leveraging the large-scale publicly available Monash time series repository~\cite{godahewa2021monash}, we curate a diverse multi-domain subset based on the anomalies-injected version released by DADA~\cite{DBLP:conf/iclr/ShentuL0SRPYG25}, comprising approximately 408 million time points for pretraining. Following \cite{zhou2023one, DBLP:conf/iclr/ShentuL0SRPYG25}, we conduct experiments on 8 popular TSAD datasets to evaluate the effectiveness of TimeRadar, covering a wide range of real-world scenarios including SMD~\cite{su2019robust}, MSL~\cite{hundman2018detecting}, PSM~ \cite{abdulaal2021practical}, SWaT~\cite{mathur2016swat}, SMAP~\cite{hundman2018detecting}, CICIDS~\cite{lai2021revisiting}, SWAN~\cite{lai2021revisiting}, and Creditcard~\cite{lai2021revisiting}. 
We also evaluate TimeRadar on the UCR benchmark~\cite{wu2021current}, with the results reported in App. \ref{app:ucr}. 
More details about the pretraining and evaluation datasets are provided in App. \ref{app:dataset}. 

\begin{table*}[!t] \footnotesize
\setlength{\tabcolsep}{8pt}
 \centering
 \caption{Quantitative results for TimeRadar (\textit{Ours}) in eight real-world datasets. The best ones are in bold, and the second ones are underlined. Avg.
denotes the average performance. 
 % The Aff-F1, AUC-R and AUC-P represent the Affiliation F1, AUC-ROC and AUC-PR (as \%) respectively.
}
 \label{tab:main}
 \newcommand{\tabincell}[2]{\begin{tabular}{@{}#1@{}}#2\end{tabular}}
 % \scalebox{1.3}{
 \begin{tabular}{c|c|c|ccccccccc}
  \toprule
  \toprule
  \textbf{Dataset} & \textbf{Setting} & \textbf{Method} & \textbf{SMD} & \textbf{MSL} & \textbf{PSM} & \textbf{SWaT} & \textbf{SMAP} & \textbf{CICIDS} & \textbf{SWAN} & \textbf{Creditcard} & \textbf{Avg.} \\
  \midrule 
  \multirow{12}{*}{\textbf{AUC-R}} & \multirow{6}{*}{\textbf{Full-shot}} & \textbf{TFMAE} & 29.37 & 45.19 & 33.35 & 56.04 & 58.49 & 50.41 & 40.21 & 47.68 & 45.09 \\
  & & \textbf{Anomaly Transformer} & 36.74 & 49.06 & 49.99 & 39.69 & 49.60 & 51.23 & 50.86 & 52.66 & 47.48 \\
  &  & \textbf{DCdetector} & 47.99 & 50.19 & 49.44 & 49.86 & 49.97 & 54.48 & 50.34 & 44.47 & 49.59 \\
  &  & \textbf{D3R} & 72.01 & 45.77 & 58.46 & 30.01 & 46.93 & 61.82 & 55.63 & 92.39 & 57.88 \\
  & & \textbf{ModernTCN} & 70.21 & 63.04 & 58.91 & 24.04 & 43.42 & 70.22 & 52.63 & 83.54 & 58.25 \\
  & & \textbf{CATCH} & 73.59 & 65.36 & \underline{64.68} & 23.49 & 47.26 & \underline{78.20} & 49.26 & 95.41 & 62.16 \\
  & & \textbf{GPT4TS} & \underline{74.15} & 58.85 & 56.26 & 23.64 & 48.19 & 64.27 & 50.88 & \underline{95.65} & 58.99 \\
  \cline{2-12}
  & \multirow{6}{*}{\textbf{Zero-shot}} & \textbf{TimesFM} & 62.80 & 69.69 & 51.04 & \underline{80.98} & 57.57 & 58.13 & 62.07 & 60.91 & 62.15 \\
  &  & \textbf{Chronos-Bolt} & 63.40 & 67.14 & 51.03 & 80.97 & \underline{58.56} & 56.45  & 62.98 & 60.85 & 62.67 \\
  &  & \textbf{Time-MoE} & 61.94 & 61.04 & 54.68 & 80.96 & 42.65 & 57.05 & \underline{63.37} & 58.99 & 60.08 \\
  &  & \textbf{SEMPO} & 57.14 & 73.46 & 51.65 & 47.22 & 51.97 & 51.99 & 46.91 & 65.27 & 55.70 \\
  &  & \textbf{DADA} & 71.98 & \underline{75.15} & 61.08 & 79.96 & 51.24 & 69.24 & 52.90 & \textbf{95.66} & \underline{69.15} \\
 & & \textbf{TimeRadar} & \textbf{74.54} & \textbf{76.02} & \textbf{67.72} & \textbf{83.31} & \textbf{59.85} & \textbf{82.76} & \textbf{81.13} & 90.41 & \textbf{76.97} \\ 
  \midrule 
  \multirow{12}{*}{\textbf{AUC-P}} & \multirow{6}{*}{\textbf{Full-shot}} & \textbf{TFMAE} & 03.15 & 09.64 & 21.54 & 13.99 & \underline{15.41} & 00.09 & 29.97 & 00.23 & 11.75 \\
  & & \textbf{Anomaly Transformer} & 03.87 & 10.47 & 28.05 & 11.71 & 12.76 & 00.10 & 33.71 & 01.81 & 12.81 \\
  & & \textbf{DCdetector} & 04.00 & 11.01 & 24.79 & 12.16 & 12.78 & 00.11 & 21.13 & 00.16 & 10.77 \\
  & & \textbf{D3R} & 14.18 & 09.45 & 38.11 & 09.45 & 11.75 & 00.14 & 42.28 & 05.17 & 16.32 \\
  & & \textbf{ModernTCN} & 14.95 & 15.20 & 38.26 & 09.21 & 10.96 & 00.14 & 45.77 & 01.65 & 17.02 \\
  & & \textbf{CATCH} & \underline{16.01} & 16.34 & \underline{43.16} & 08.70 & 12.14 & \underline{00.20} & 42.96 & 09.08 & 18.57 \\
  & & \textbf{GPT4TS} & 15.85 & 13.83 & 35.82 & 08.41 & 12.23 & 00.12 & 46.00 & 08.30 & 17.57 \\
  \cline{2-12}
  & \multirow{6}{*}{\textbf{Zero-shot}} & \textbf{TimesFM} & 07.37 & 20.26 & 29.30 & \underline{69.32} & 14.31 & 00.14 & 50.24 & 00.28 & 23.90 \\
  &  & \textbf{Chronos-Bolt} & 07.64 & 18.72 & 29.24 & 69.19 & 15.25 & 00.13 & 49.95 & 00.32 & 23.80 \\
  &  & \textbf{Time-MoE} & 07.25 & 16.79 & 32.31 & 69.21 & 11.62 & 00.14 & \underline{50.92} & 00.26 & 23.56 \\    
  &  & \textbf{SEMPO} & 06.60 & 22.04 & 30.58 & 10.93 & 12.55 & 00.12 & 31.17 & 01.74 & 14.97 \\
  &  & \textbf{DADA} & 13.16 & \textbf{26.42} & 38.79 & 50.59 & 12.08 & 00.14 & 48.63 & \underline{09.10} & \underline{24.86} \\
  &  & \textbf{TimeRadar} & \textbf{16.09} & \underline{24.38} & \textbf{44.44} & \textbf{70.33} & \textbf{16.42} & \textbf{00.38} & \textbf{76.02} & \textbf{09.32} & \textbf{32.67} \\ 
      \midrule  
  \multirow{12}{*}{\textbf{Aff-F1}} & \multirow{6}{*}{\textbf{Full-shot}} &
  \textbf{TFMAE} & 63.36 & 67.04 & 65.64 & 68.89 & 70.72 & 25.34 & 09.19 & 41.90 & 51.51 \\
  & & \textbf{Anomaly Transformer} & 66.42 & 66.49 & 65.37 & 69.39 & 71.65 & 34.70 & 33.67 & 62.81 & 58.81 \\
  & & \textbf{DCdetector} & 66.45 & 70.56 & 66.99 & 69.03 & 68.99 & 45.02 & 24.01 & 59.65 & 58.84 \\
  & & \textbf{D3R} & 78.02 & 77.02 & 80.29 & \underline{74.39} & 74.09 & 67.79 & 43.19 & 70.40 & 70.65 \\
  & & \textbf{ModernTCN} & 83.16 & 77.17 & 79.59 & 71.13 & 67.41 & 51.74  & 46.45 & 73.80 & 68.81 \\
  & & \textbf{CATCH} & 84.46 & 70.97 & 79.50 & 71.82 & 66.06 & 72.66 & 53.37 & \underline{73.85} & 71.59 \\
  & & \textbf{GPT4TS} & 83.13 & 77.23 & 81.44 & 70.11 & \underline{74.67} & 54.00 & 47.27 & 72.88 & 70.09 \\
  \cline{2-12}
  & \multirow{6}{*}{\textbf{Zero-shot}} & \textbf{TimesFM} & 72.52 & 74.17 & 69.58 & 71.79 & 72.00 & 67.45 & 69.06 & 68.30 & 70.61 \\
  & & \textbf{Chronos-Bolt} & 68.34 & 75.78 & 69.14 & 70.98 & 73.91 & 67.41 & 68.82 & 67.55 & 70.24 \\
  &  & \textbf{Time-MoE} & 72.23 & 66.05 & 69.02 & 69.02 & 70.87 & 64.39 & 62.22 & 66.43 & 67.53 \\
  &  & \textbf{SEMPO} & 72.35 & 74.45 & 73.07 & 60.31 & 66.18 & 34.35 & 71.25 & 68.07 & 64.88 \\
  & & \textbf{DADA} & \underline{84.57} & \underline{78.27} & \underline{81.58} & 73.70 & 74.13 & \underline{73.49} & \textbf{71.93} & \textbf{74.84} & \underline{76.56} \\
  & & \textbf{TimeRadar} & \textbf{84.80} & \textbf{79.44} & \textbf{82.03} & \textbf{76.73} & \textbf{75.41} & \textbf{73.51} & \underline{71.06} & 72.34 & \textbf{76.92} \\ 
  \bottomrule 
    \midrule 
 \end{tabular}
 % }
\end{table*}

\noindent \textbf{Competing methods.}
% TimeRadar is mainly evaluated against 13 baseline models for comprehensive comparison, 
% covering approaches: (1) \textbf{Time-domain based methods}: Anomaly Transformer~\cite{xu2021anomaly}, DCdetector~\cite{yang2023dcdetector}, D3R~ \cite{wang2023drift}, GPT4TS~\cite{zhou2023one}, ModernTCN~\cite{luo2024moderntcn}, Time-MoE \cite{DBLP:conf/iclr/ShiWNLYWJ25}, Chronos-Bolt~\cite{ansari2024chronos}, and one generalist TSAD model DADA~\cite{DBLP:conf/iclr/ShentuL0SRPYG25}; (2) Frequency domain methods, including 
%  SEMPO~\cite{he2025sempo} and CATCH~\cite{DBLP:conf/iclr/WuQL0HGXY25}; (3) Time-frequency domain method TFMAE \cite{fang2024temporal}. 
 TimeRadar is evaluated against 13 state-of-the-art (SOTA) methods for comprehensive comparison, including (1) \textit{Pretrained TSFMs}: GPT4TS~\cite{zhou2023one}, TimesFM~\cite{DBLP:conf/icml/DasKSZ24}, Chronos-Bolt~\cite{ansari2024chronos}, Time-MoE~\cite{DBLP:conf/iclr/ShiWNLYWJ25}, SEMPO~\cite{he2025sempo}, and DADA~\cite{DBLP:conf/iclr/ShentuL0SRPYG25}, where DADA is a generalist model for TSAD; (2) \textit{Conventional TSAD models}: TFMAE~\cite{fang2024temporal}, Anomaly Transformer~\cite{xu2021anomaly}, DCdetector~\cite{yang2023dcdetector}, D3R~\cite{wang2023drift}, ModernTCN~\cite{luo2024moderntcn}, CATCH~\cite{DBLP:conf/iclr/WuQL0HGXY25}. In terms of modeling domains, SEMPO and CATCH operate in the frequency domain, TFMAE in the time-frequency domain, while the remaining models work in the time domain. 
 Detailed descriptions of of these baselines are provided in App. \ref{app:competing_methods}.
We adopt the default settings for all baselines, including evaluation metrics and thresholding protocols, following common practices in anomaly detection to ensure a fair comparison. 
% Note that some task-specific deep models can not be adapted to the zero-shot setting. For these baselines, we instead report their full-shot performance evaluated on the test dataset. 
Some earlier baseline methods are also considered, with the detailed comparison results provided in App. \ref{app:additional_quantitative}. 

% We also evaluate the TimeRadar with 10 strong baseline methods on the NeurIPS-TS benchmark \cite{lai2021revisiting}, and the experimental results are shown in the Table. \ref{table 2}. 
% Affiliation results comparing with more full-shot baselines are provided in the Appendix Table \ref{table 2}.

\noindent \textbf{Evaluation metrics.}
During the evaluation, we deliberately avoid using Point Adjustments (PA) to adjust the detection result, given that some works\cite{huet2022local} have demonstrated that PA can lead to faulty performance evaluations. Following the previous work \cite{DBLP:conf/iclr/NieNSK23, lan2025towards}, we use the widely-used affiliation-based F1-score (Aff-F1) \cite{huet2022local}, recently 
\cite{wang2023drift, yang2023dcdetector}, which can take into account the average directed distance between predicted anomalies and ground truth anomalies. We also employed two standard metrics, including AUC-R and AUC-P, which evaluate the detection performance in terms of the area under the ROC curve and the Precision–Recall curve, respectively. More auxiliary metrics for TSAD evaluation are also reported, with the results provided in App.~\ref{app:additional_metric}.

\noindent \textbf{Implementation details.}
The entire pre-training process takes 3 hours on 2 Pro6000-96G GPUs with BF32 precision and a batch size of 2,048. By default, we set window size $T=100$, patch size $T_p=5$, hidden dimension $D_p=64$, output dimension of $\mathrm{Conv1D}$ $D_o=256$, contextual deviation margin $\gamma=2$. The weight parameter $\lambda$ controlling the weight of CDL is set to 0.1. 
The hyperparameter analysis of $T$, $\gamma$, and $\lambda$ is given in App.~\ref{app:additional_sensitive}.
Regarding optimization, we employ the AdamW optimizer with hyperparameters: learning rate of 1e-3, weight decay of 0.1, momentum coefficients $\beta_1=0.9$ and $\beta_2=0.95$. The learning rate is linearly warmed up during the first 10,000 steps and kept constant thereafter.
% More implementation details are provided in Appendix. 

\subsection{Zero-shot Anomaly Detection}
As GPT4TS~\cite{zhou2023one} and conventional TSAD models such as Anomaly Transformer~\cite{xu2021anomaly} are not designed for zero-shot transfer, they are trained separately on the training split of the target dataset and evaluated on its test split, \ie, under the full-shot setting.
In contrast, TSFMs follow a zero-shot protocol, being pretrained on multi-domain datasets and directly evaluated on downstream targets without any additional fine-tuning. Table \ref{tab:main} reports the zero-shot performance of FMs and the full-shot performance of conventional TSAD models and GPT4TS.

From the results, we have three key observations. (1) TimeRadar consistently surpasses the zero-shot TSAD detector DADA on the majority of datasets, achieving SOTA performance across AUC-R, AUC-P, and Aff-F1. In particular, TimeRadar yields average improvements of 10.5\% in AUC-R and 29.4\% in AUC-P.
% , demonstrating pronounced advantages in detection quality. 
This performance gain is largely attributed to the adaptive rotation of the input series into a unified continuous time-frequency domain with a learnable, input-specific angle. Such a transformation effectively exposes anomalies that may remain subtle or indistinguishable in either the pure time or frequency domain, thereby enhancing their saliency and detectability while mitigating cross-domain mismatch. 
This advantage is further evidenced by the fact that TimeRadar surpasses the pure frequency-domain models CATCH and SEMPO, as well as the time–frequency model TFMAE.
% In contrast to DADA, which reconstructs features in the time domain, it failed to explicitly address the cross-domain conflicts. 
(2) FMs such as TimesFM, Chronos-Bolt, Time-MoE, and SEMPO 
are primarily designed for time series forecasting rather than anomaly detection, and therefore struggle to capture domain-invariant and discriminative normality and abnormality, resulting in inferior cross-domain performance;
(3) Generalist TSAD models, \ie, TimeRadar and DADA, surpass conventional TSAD baselines that are trained and evaluated on a single dataset. This strongly suggests that large-scale pretraining on diverse datasets yields far more transferable notions of normality and abnormality, while single-dataset training is inherently limited.

% Compared with existing FMs such as DADA and SEMPO, TimeRadar also has faster inference and contains only 1.6M parameters, achieving a better trade-off between performance and efficiency than both conventional deep models and FMs, see details in App.~\ref{app:efficiency}.

% , improving performance when transferred to new domains. 
% The general baseline failed to deliver strong performance across all datasets because it was trained exclusively on a single specific time series dataset. 
% (3) As a unified time–frequency domain model, TimeRadar surpasses the pure frequency-domain models CATCH and SEMPO, as well as the time–frequency model TFMAE. The main reason is that single-domain and fusion-based time–frequency models are primarily designed for conventional TSAD settings and do not explicitly learn domain-invariant and discriminative representations.

\subsection{Few-shot Anomaly Detection}
\begin{figure} [!t]
 \centering
 \includegraphics[width=0.46\textwidth]{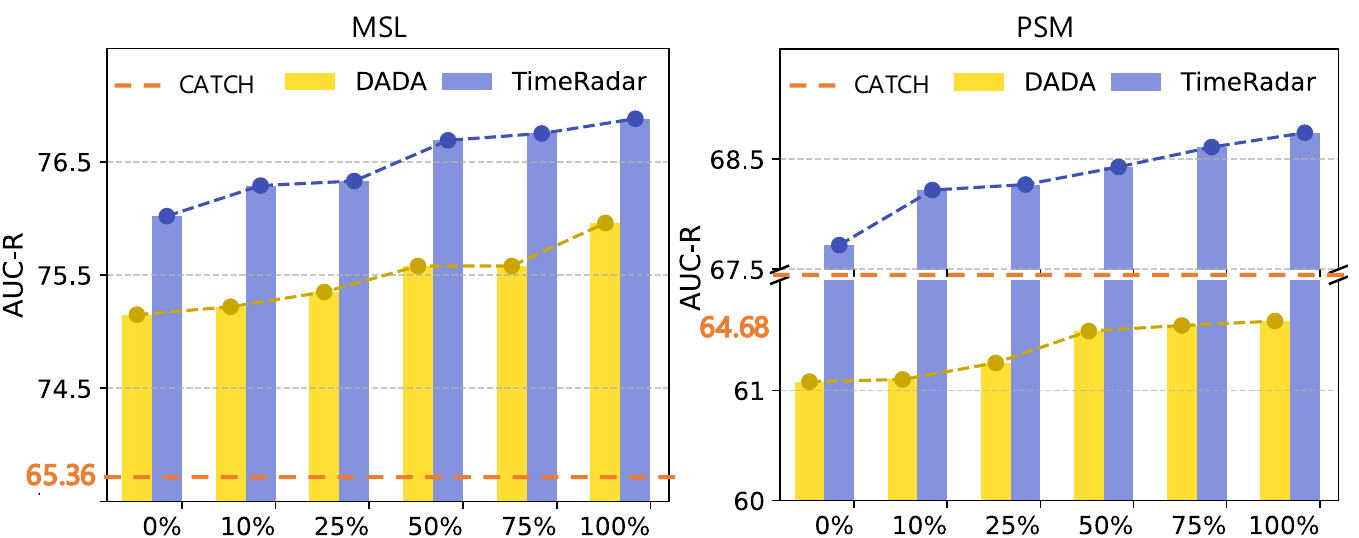}
 \caption{Few-shot performance of TimeRadar and DADA using different amount of training data in a target dataset.
 % , along with the strongest full-shot baseline CATCH.
 }
 \label{fig:fine_tuning}
 % \vspace{-1em}
\end{figure}

As shown in Fig. \ref{fig:fine_tuning}, following DADA \cite{DBLP:conf/iclr/ShentuL0SRPYG25}, we further fine-tune TimeRadar and DADA on two downstream datasets, MSL and PSM, under varying levels of data scarcity to facilitate domain-specific adaptation. The performance of both models improves as the proportion of available training data increases. Notably, TimeRadar consistently outperforms DADA across all training-data percentages and even surpasses the strongest full-shot baseline, CATCH. These findings demonstrate the pivotal role of TimeRadar’s pretrained initialization, which equips the model with strong inductive biases for downstream learning. Consequently, TimeRadar achieves more effective adaptation to domain-specific patterns even in resource-constrained regimes and exhibits a stronger ability to transfer knowledge to previously unseen data.

\begin{table*}[!t] \footnotesize
\setlength{\tabcolsep}{0.9pt}
 \centering
 \caption{Ablation study results. The higher values indicate better performance. The best ones are in bold. 
}
 \label{tab:ablation}
 \newcommand{\tabincell}[2]{\begin{tabular}{@{}#1@{}}#2\end{tabular}}
 % \scalebox{0.97}{
 \begin{tabular}{c|cccccccccccccccccc}
  \toprule
  \toprule
  \multirow{2}{*}{\textbf{Design}} & \multicolumn{3}{c}{\textbf{MSL}} & \multicolumn{3}{c}{\textbf{SMAP}} & \multicolumn{3}{c}{\textbf{SWaT}} & \multicolumn{3}{c}{\textbf{PSM}} & \multicolumn{3}{c}{\textbf{CICIDS}} & \multicolumn{3}{c}{\textbf{SWAN}} \\
  \cmidrule(r){2-4} \cmidrule(r){5-7} \cmidrule(r){8-10} \cmidrule(r){11-13} \cmidrule(r){14-16} \cmidrule(r){17-19}
  & \textbf{Aff-F1} & \textbf{AUC-R} & \textbf{AUC-P} & \textbf{Aff-F1} & \textbf{AUC-R} & \textbf{AUC-P} & \textbf{Aff-F1} & \textbf{AUC-R} & \textbf{AUC-P} & \textbf{Aff-F1} & \textbf{AUC-R} & \textbf{AUC-P} & \textbf{Aff-F1} & \textbf{AUC-R} & \textbf{AUC-P} & \textbf{Aff-F1} & \textbf{AUC-R} & \textbf{AUC-P}  \\ 
  \midrule 
   TimeRadar & \textbf{79.44} & \textbf{76.02} & \textbf{24.38} & \textbf{75.41} & \textbf{59.85} & \textbf{16.42} & \textbf{76.73} & \textbf{83.31} & \textbf{70.33} & \textbf{82.03} & \textbf{67.72} & \textbf{44.44} & \textbf{73.51} & \textbf{82.76} & \textbf{00.38} & \textbf{71.06} & \textbf{81.13} & \textbf{76.02} \\
   \midrule 
  \textbf{A.1} Frequency domain + CDL & 76.69 & 72.61 & 22.57 & 73.74 & 58.52 & 15.10 & 73.46 & 80.23 & 69.33 & 79.20 & 63.98 & 38.99 & 69.48 & 78.64 & 00.24 & 68.37 & 77.39 & 72.22 \\ 
  \textbf{A.2} Time domain + CDL & 70.03 & 73.06 & 21.49 & 70.67 & 51.32 & 12.31 & 70.52 & 82.27 & 69.72 & 71.79 & 65.45 & 44.29 & 67.96 & 73.30 & 00.20 & 70.38 & 79.39 & 69.04 \\
  \midrule 
  \textbf{B} BottleneckCNN Encoder & 77.66 & 75.08 & 23.84 & 73.27 & 58.26 & 14.96 & 74.63 & 80.46 & 69.38 & 80.12 & 64.96 & 40.73 & 70.12 & 79.20 & 00.25 & 69.41 & 76.92 & 72.21 \\
\midrule 
  \makecell{\textbf{C} TimeRadar(w/o CDL)} & 77.33 & 75.40 & 23.86 & 74.03 & 58.81 & 15.73 & 74.51 & 80.39 & 69.60 & 79.47 & 62.02 & 38.72 & 69.26 & 80.73 & 00.27 & 68.19 & 78.43 & 73.74 \\
  \bottomrule 
  \bottomrule
 \end{tabular}
 % }
\end{table*}

\begin{figure*}[!t] 
\centering
\subfloat[PSM]{
    \label{fig9a}
    \includegraphics[height=2.9cm]{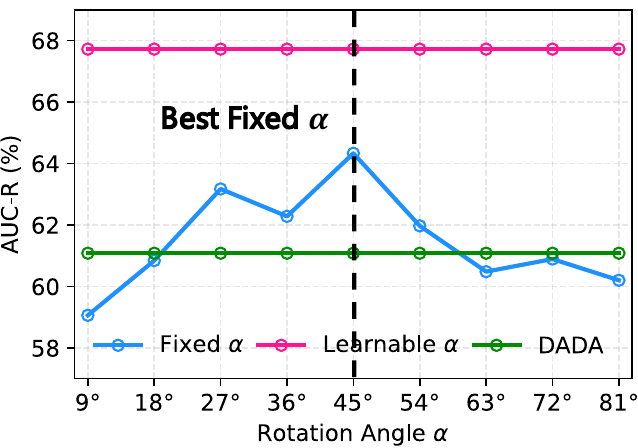}
    }  
\subfloat[SWaT]{
    \label{fig9b}
    \includegraphics[height=2.9cm]{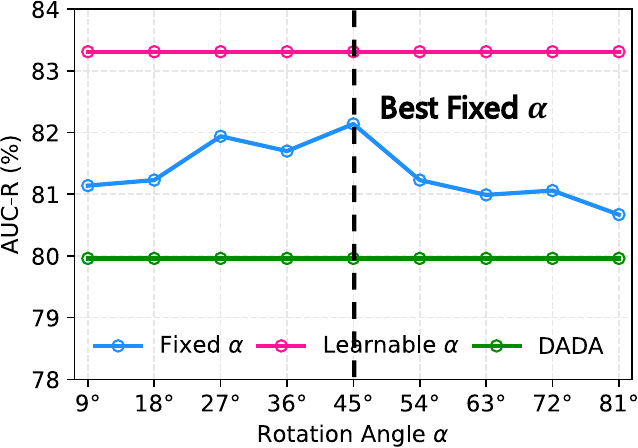}
    }
\subfloat[MSL]{
    \label{fig9c}
    \includegraphics[height=2.9cm]{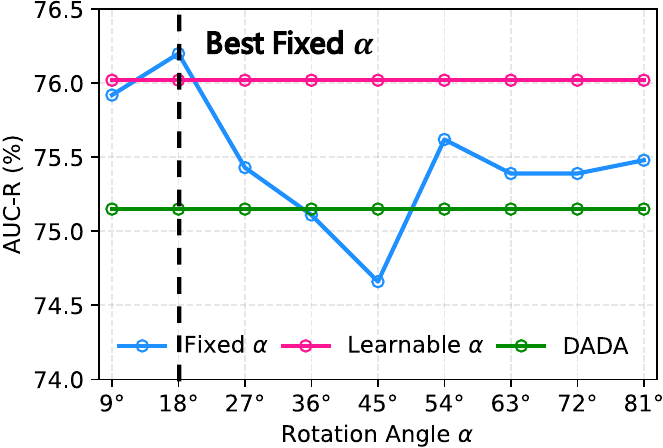}
    }
\subfloat[SMD]{
    \label{fig9d}
    \includegraphics[height=2.9cm]{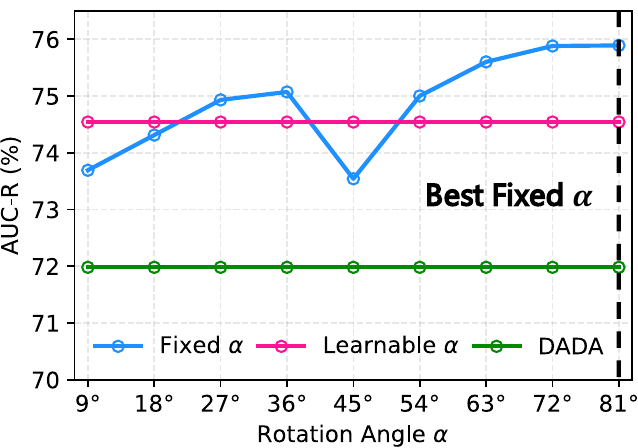}
    }
\centering
\caption{Comparison of TimeRadar (learnable rotation angle) to its variants with manually searched rotation angles.}
\label{fig:rotation}
\end{figure*}

\subsection{Ablation Study} \label{sec:ablation}
\noindent\textbf{Analysis of key components.} 
To verify the effectiveness of each component in TimeRadar, we designed four variants, including: \textbf{A.1 Frequency domain + CDL}, which replaces the FRFT with the FT and employs only the BottleneckCNN within \textit{FracBlocks} as the encoder; \textbf{A.2 Time domain + CDL}, which removes the FRFT, adopts the BottleneckCNN as the encoder, and utilizes a real-valued MLP as the reconstruction head; \textbf{B BottleneckCNN Encoder}, which uses the BottleneckCNN in \textit{FracBlocks} as the encoder without fractional modulation; and \textbf{C TimeRadar (w/o CDL)}, which removes the CDL module during pretraining.
The experimental results are summarized in Table~\ref{tab:ablation}. We observe that introducing CDL in either the time domain \textbf{A.2} or frequency domain \textbf{A.1} in isolation yields inferior performance compared with TimeRadar. This underscores the necessity of a unified and continuous time–frequency transformation, which enables learning in a more discriminative and expressive representation space. By adaptively aligning the time–frequency characteristics of normal patterns, TimeRadar better separates normal from abnormal patterns, thereby exposing anomalies that may remain subtle or indistinguishable when modeled in a single domain.
% and improving detectability on unseen data.
Furthermore, the variant \textbf{C} consistently degrades performance, corroborating the crucial role of CDL in enhancing normal-abnormal separability in the time-frequency representation space by encouraging compact normal clusters and enforcing a clear margin from anomalies. 
Finally, removing fractional modulation from \textit{FracBlocks}, \ie, the variant \textbf{B}, also reduces performance, justifying the importance of local phase alignment across patches in the time–frequency domain, which makes the learned representations more compatible with convolutional weight sharing. 
% More comparative analyses between TimeRadar with a learnable rotation angle and its variant with different fixed rotation angles are given in App.~\ref{app:additional_rotation}. 

\noindent\textbf{Effectiveness of learnable fractional order.}
To further demonstrate the effectiveness of TimeRadar with a rotatable domain, we compare it with variants using a wide range of various rotation angles $\alpha$ on four datasets, with a step size of $9\degree$. The results are shown in Fig.~\ref{fig:rotation}, with DADA's performance used as baseline. It is clear that (1) the best rotation angle often differ largely in different datasets 
and (2) 
% can improve TimeRadar’s performance, but the gain is highly sensitive to the chosen angle value. In contrast, introducing a
the rotation angle learned by TimeRadar yields highly comparable performance to the best rotation angle found in the grid search across all four datasets, 
% and enables TimeRadar to outperform the strong baseline model DADA. This is mainly because
demonstrating that the learned rotation angle in TimeRadar can effectively and adaptively align each dataset to a dataset-specific orientation in the continuous time–frequency space. This offers TimeRader the strong capability of flexibly adapting to a new dataset based on its data input, supporting its superior generalizability. 
% while still learning generalizable and domain-robust features. 

\subsection{Further Analysis of TimeRadar}
\begin{figure}[!t]
 \centering
 \includegraphics[width=0.48\textwidth]{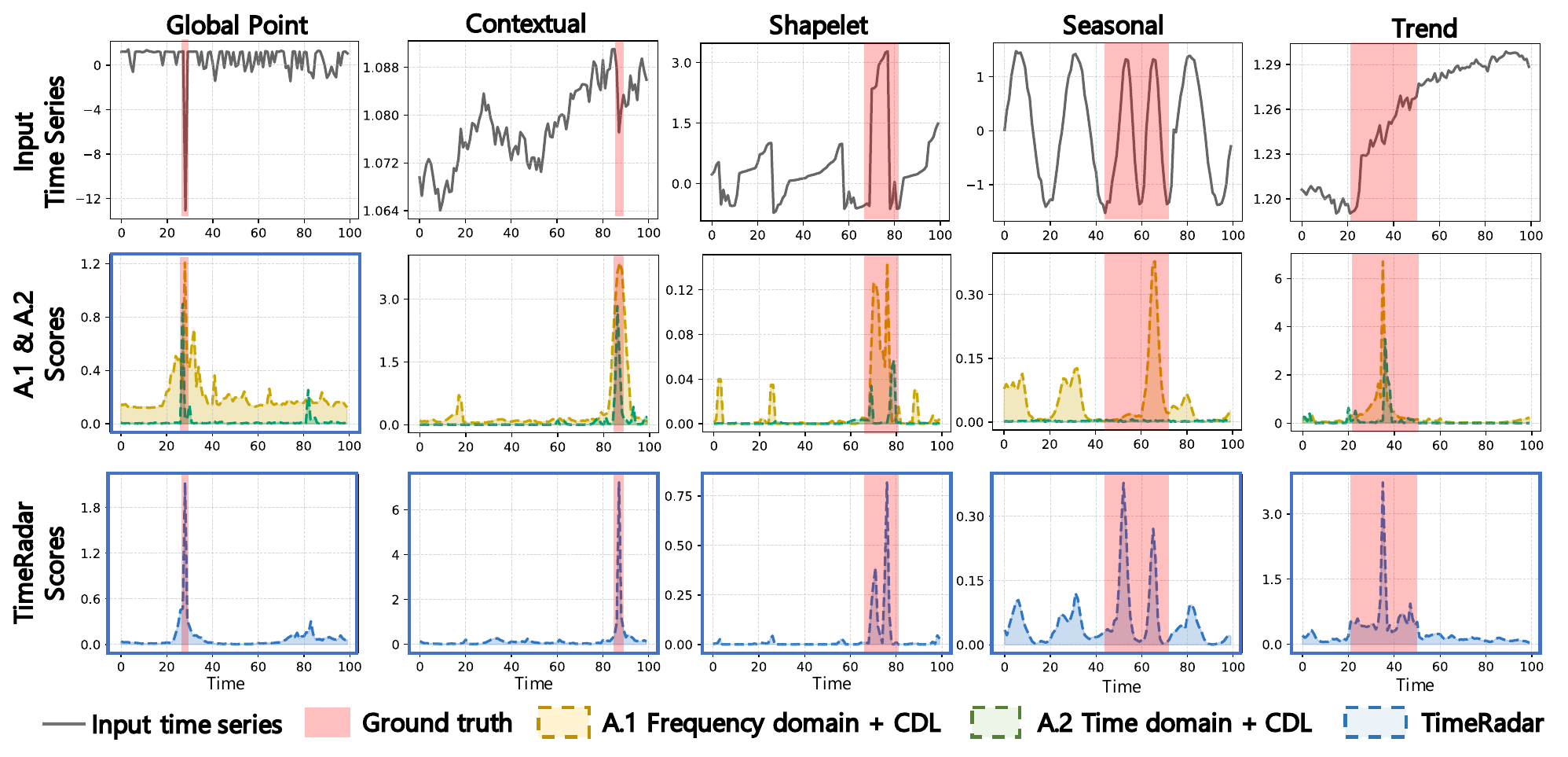}
 \caption{
Visualization of anomaly scores.
% produced by TimeRadar and variants for different categories of anomalies. 
% Time points whose anomaly scores exceed the threshold are classified as anomalous. 
% The outer blue bounding box indicates that anomalies of this type can be identified more effectively.
}
 \label{fig:visualization}
 \vspace{-1em}
\end{figure}

\noindent \textbf{Visualization of anomaly scores.}
To further demonstrate the effectiveness of adaptively transforming the series into an data-dependent continuous time-frequency domain, we conduct a comparative visualization of anomaly scores produced by TimeRadar and its variants, \textbf{A.1 Frequency domain + CDL} and \textbf{A.2 Time domain + CDL}, as discribed in Sec.~\ref{sec:ablation}. As shown in Fig. \ref{fig:visualization}, we present five representative anomaly types~\cite{lai2021revisiting} spanning point and subsequence anomalies, where all cases are drawn from real-world datasets except for the seasonal anomaly from synthetic datasets~\cite{DBLP:conf/iclr/WuQL0HGXY25}. Successful detection cases are highlighted by blue boxes.
\begin{figure}
  \centering
  \includegraphics[width=0.4\textwidth]{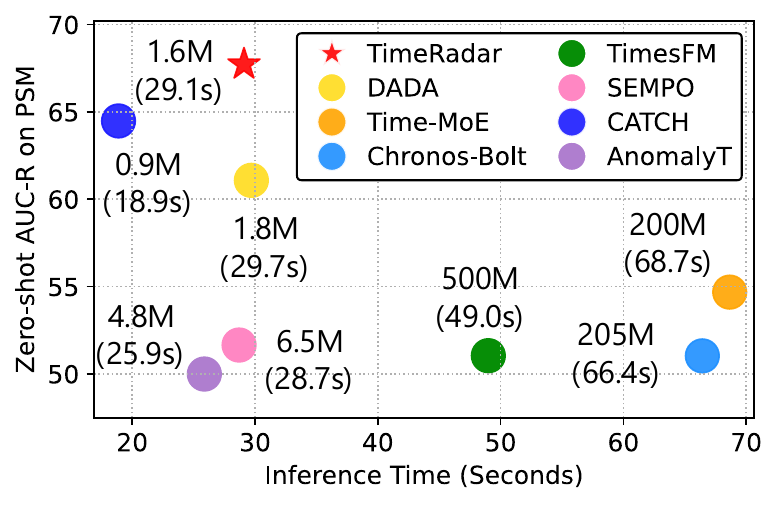}
  \caption{Efficiency comparison on PSM.}
  \label{fig:efficiency}
  \vspace{-1em}
\end{figure}
% \begin{figure}[!t]
%  \centering
%  % Requires \usepackage{graphicx}
%  \includegraphics[width=0.48\textwidth]{figures/fig9e.pdf}
%  \caption{Sensitivity analysis w.r.t weight parameter $\lambda$.
% }
%  \label{fig:weight}
% \end{figure}
From the results, we can observe that for simple point anomalies (first and second columns), global point anomalies are consistently identified by all three methods, whereas the frequency-domain variant \textbf{A.1} fails to correctly detect the contextual point anomalies.
For more complex subsequence anomalies (third to fifth columns), the variant \textbf{A.1} shows strong sensitivity to the true boundaries of shapelet and seasonal anomalies, while the variant \textbf{A.2} exhibits either excessively sharp responses (\eg, shapelet and trend anomalies) or overly smooth variations (\eg, seasonal anomalies) around the anomaly regions, leading to unreliable recognition. This observation aligns with the findings in Fig. \ref{fig:example}. In contrast, TimeRadar accurately identifies all challenging subsequence anomalies, benefiting from the learnable fractional time-frequency domain.
% Overall, these visualizations demonstrate TimeRadar's strong capability in detecting challenging anomalies.

\noindent \textbf{Efficiency analysis.} \label{app:efficiency}
% To further highlight the performance and efficiency advantages, 
We compare the inference costs
of TimeRadar with five FMs as well as two conventional anomaly detectors, CATCH and Anomaly Transformer, on the PSM dataset, as shown in Fig.\ref{fig:efficiency}. All experiments are conducted on a single Pro6000 GPU with a batch size of 128.
Compared with existing FMs such as DADA, Chronos-bolt and TimesFM, TimeRadar has faster inference and contains only 1.6M parameters. Although its parameter count and inference time are higher than those of conventional TSAD models, TimeRadar provides a more favorable trade-off between effectiveness and efficiency than both conventional TSAD models and large-scale TSFMs, making it more practical for real-world deployment.

\section{Conclusion}
In this paper, we introduce a novel FM, TimeRadar, which is the first TSFM that learns to adaptively detect diverse anomalies across different datasets in a domain-rotatable time–frequency space. TimeRadar is implemented with FTFRecon and CDL. FTFRecon enables input-dependent data reconstruction in an optimal time–frequency representation by adaptively rotating each time series to the most pronounced angle between the continuous time and frequency domain. CDL is designed to capture local abnormalities by modeling how an input segment deviates from its surrounding context in the learned fractional time–frequency domain. Extensive experiments show the superiority of TimeRadar over SOTA conventional TSAD and TSFM methods under both zero-shot and few-shot settings.
% generalizes strongly to unseen datasets, consistently delivering the best results compared with generalist detectors and conventional dataset-specific models.

%\begin{acks}
%To Robert, for the bagels and explaining CMYK and color spaces.
%\end{acks}

\bibliographystyle{ACM-Reference-Format}
\bibliography{sample-base}

\appendix

\section{Experimental Details}

\subsection{Datasets}  \label{app:dataset}
\noindent \textbf{Pretraining datasets.}
The emergence of large-scale publicly available time series collections, \eg, Anomaly Detection Data~\cite{DBLP:conf/iclr/ShentuL0SRPYG25}, Monash$^+$~\cite{DBLP:conf/iclr/ShentuL0SRPYG25}, and TSB-AD~\cite{lan2025towards}, have resolved data availability issues for building generalist time series anomaly detection models. Theses collections contain both normal and abnormal time series spanning diverse domains, reflecting a wide spectrum of real-world applications. 
In this work, we leverage Monash$^+$ as our pretraining dataset, comprising approximately 408 million time points. The normal time series in Monash$^+$ originates from the Monash Prediction Library~\cite{godahewa2021monash}, whose key properties, including domain, sampling frequency, and sequence length are summarized in Table \ref{tab:pre_training}. The abnormal time series is synthesized by injecting anomalies into the Monash normal series. Importantly, none of
the downstream evaluation datasets are included in our pretraining data. We then set the pretraining training-validation split to 8:2. In the CDL module, patch-wise labels follow a coarse-grained assignment strategy, where a patch is labeled as anomalous if it contains at least one anomalous time point, and otherwise it is labeled as normal.

Following DADA~\cite{DBLP:conf/iclr/ShentuL0SRPYG25}, the anomaly injection procedure is summarized as follows. A subsequence from the original time series and an anomaly type are randomly sampled. If the selected subsequence has not been previously modified, the chosen anomaly is injected into it. This process is repeated until the overall anomaly ratio of the sequence reaches a predefined target level.

\begin{table}[h]
  \caption{Multi-domain time series datasets from the Monash Prediction Library \cite{godahewa2021monash}.}
   \label{tab:pre_training}
\begin{tabular}{cccc}
\hline \hline
\textbf{Dataset} & {\textbf{Domain}} & \textbf{Frequency} & {\textbf{Length}} \\ \hline
Aus. E.D. & Energy & Half Hourly & 1,155,264 \\
Wind  & Energy & 4 Seconds & 7,397,147 \\
Wind Farms & Energy & Minutely & 172,178,060 \\
Solar Power & Energy & 4 Seconds & 7,397,223\\
Solar & Energy & 10 Minutes & 7,200,857\\
London Smart M. & Energy & Half Hourly & 166,527,216 \\
Saugeen River F. & Nature & Daily & 23,741 \\
Sunspot & Nature & Daily & 73,924  \\
Weather & Nature & Daily & 43,032,000  \\
KDD Cup 2018 & Nature & Daily & 2,942,364 \\
US Births & Nature & Daily & 7,305   \\
FRED\_MD & Economic & Monthly & 77,896   \\
Bitcoin & Economic & Daily & 75,364  \\
NN5   & Economic & Daily & 87,801 \\
\hline \hline
\end{tabular}
\end{table}

\begin{table*}[!t]
\setlength{\tabcolsep}{10pt}
 \centering
  \caption{The datasets used for evaluation, where AR represents the anomaly ratio.}
   \label{tab:evaluation}
\begin{tabular}{ccccccc}
\hline \hline
\textbf{Dataset} & {\textbf{Domain}} & {\textbf{Dimension}}    & {\textbf{Training}} & {\textbf{Validation}} & {\textbf{Test(labeled)}} & {\textbf{AR(\%)}}\\ \hline

MSL & Spacecraft & 1 & 46,653 & 11,664 & 73,729 & 10.5\\
PSM & Server Machine & 25 & 105,984 & 26,497 & 87,841 & 27.8\\
SMAP & Spacecraft & 1 & 108,146 & 27,037 & 427,617 & 12.8\\
SMD & Server Machine & 38 & 566,724 & 141,681 & 708,420 & 4.2\\
SWaT & Water treatment & 31 & 396,000 & 99,000 & 449,919 & 12.1\\
Creditcard & Finance & 29 & 113,922 & 28,481 & 142,404 & 0.17\\
% GECCO & Water treatment & 9 & 55,408 & 13,852 & 69,261 & 1.25\\
CICIDS & Web & 72 & 68,092 & 17,023 & 85,116 & 1.28\\
SWAN & Space Weather & 38 & 48,000 & 12,000 & 60,000 & 23.8\\
UCR & Natural & 1 & 1,790,680 & 447,670 & 6,143,541 & 0.6 \\
\hline \hline
\end{tabular}
\end{table*}

\noindent \textbf{Evaluation datasets.}
For zero- and few-shot anomaly detection, we evaluate TimeRadar and the baselines on a diverse collection of datasets spanning multiple domains, including spacecraft, servers, water treatment, finance, networking, and space weather. The datasets include SMD~\cite{su2019robust}, MSL~\cite{hundman2018detecting}, PSM~ \cite{abdulaal2021practical}, SWaT~\cite{mathur2016swat}, SMAP~\cite{hundman2018detecting}, CICIDS~\cite{lai2021revisiting}, SWAN~\cite{lai2021revisiting},  Creditcard~\cite{lai2021revisiting}, as well as the UCR benchmark~\cite{wu2021current}. The UCR benchmark contains 250 sub-datasets, each consisting of a univariate series with a single anomaly segment. Following previous work~\cite{DBLP:conf/iclr/ShentuL0SRPYG25}, only the first continuous dimension is retained for MSL and SMAP~\cite{wang2023drift}. A detailed statistical summary of the datasets is provided in Table~\ref{tab:evaluation}. The feature dimensions ranges from 1 to 72, sequence lengths vary between 58,317 and 708,405, and anomaly ratios span from 0.17\% to 27.8\%. Such diversity reflects substantial heterogeneity across datasets and enables a comprehensive evaluation of TimeRadar.

 \begin{table*}[!t]
 \centering
 \caption{Comparison between pre-trained FMs on time series.}
 \label{table_pre-trained FMs}
 \newcommand{\tabincell}[2]{\begin{tabular}{@{}#1@{}}#2\end{tabular}}
 \scalebox{0.97}{
 \begin{tabular}{ccccccc}
  \toprule
  \toprule
  Method & TimeRadar & DADA & SEMPO & Time-MoE & Chronos-Bolt & TimesFM  \\
  \midrule
  Architecture & Encoder-only & Encoder-decoder & Encoder-only & Decoder-only & Encoder-decoder & Decoder-only\\
  \midrule
  Model Size & 1.6M & 1.8M & 6.5M & 200M & 205M & 500M \\
  \midrule
  Model version & - & - & Small & Large & Base & v2.0 \\
  \midrule
  Input Token & Patch & Patch & Patch & Point & Point & Patch \\
  \midrule
  Dataset Scale & 408M & 473M & 83M & 309B & 84B & 100B \\
  \midrule
  Context Length & 100 & 100 & 512 & $\leq$ 4096 & $\leq$ 512&$\leq$ 512\\
  \midrule
  Source & Ours & ~\cite{DBLP:conf/iclr/ShentuL0SRPYG25} & ~\cite{he2025sempo} & ~\cite{DBLP:conf/iclr/ShiWNLYWJ25} & ~\cite{ansari2024chronos} & \cite{DBLP:conf/icml/DasKSZ24} \\
  \bottomrule
  \bottomrule
 \end{tabular}
}
\end{table*}

\subsection{Competing Baselines} 
\label{app:competing_methods}

\noindent \textbf{Full-shot competing baselines.} The model is trained separately on the target dataset’s training set and evaluated on its corresponding test set.
 \begin{itemize}

    \item  TFMAE~\cite{fang2024temporal} is a Temporal-Frequency Masked AutoEncoder that utilizes a contrastive criterion to detect anomalies in time series. It contains two Transformer-based autoencoders that respectively incorporate a window-based temporal masking strategy and an amplitude-based frequency masking strategy to learn knowledge without abnormal bias and reconstruct anomalies using the extracted normal information. 
      
  \item Anomaly Transformer~\cite{xu2021anomaly} is a new anomaly-Attention mechanism to compute the association discrepancy, where a minimax strategy is devised to amplify the normal-abnormal distinguishability of the association discrepancy.

      \item DCdetector~\cite{yang2023dcdetector} is a multi-scale
dual attention contrastive representation learning model by utilizing a novel dual attention asymmetric. It is designed to create the permutated environment and pure contrastive loss to learn a permutation-invariant representation with superior discrimination abilities.

        \item D3R~\cite{wang2023drift} tackles the drift via decomposition and reconstruction. In the decomposition, it utilizes data-time mix-attention to dynamically decompose long-period multivariate time series, overcoming the limitation of the local sliding window. The noise diffusion and directly reconstructing the polluted data are employed to avoid retraining once the bottleneck changes.
 
      \item CATCH~\cite{DBLP:conf/iclr/WuQL0HGXY25} is a framework based on frequency patching, where a Channel Fusion Module (CFM) is encouraged to iteratively discover appropriate patch-wise channel correlations. It clusters relevant channels while isolating adverse effects from irrelevant channels by optimizing a bi-level multi-objective.
       \item GPT4TS~\cite{zhou2023one} leverages the pre-trained language model and fine-tunes the input embedding, positional embeddings, layer normalization, and output layer to make the pre-trained model adapt to time series data. 
        \item ModernTCN~\cite{luo2024moderntcn} conducts some time series-related modifications to the traditional TCN by using a modern pure convolution structure. It aims to efficiently utilize cross-time and cross-variable dependency for general time series analysis.
 \end{itemize}

\noindent \textbf{Zero-shot competing baselines.} The model is trained on multiple source-domain datasets and then directly evaluated on the downstream target-domain test sets without any additional fine-tuning. A comparison
between these pre-trained FMs is presented in Table~\ref{table_pre-trained FMs}.
 \begin{itemize}
     \item TimesFM \cite{DBLP:conf/icml/DasKSZ24}  introduces input patching to represent sequences efficiently, leverages longer output patches to reduce autoregressive steps, and employs random masking to learn from variable-length contexts.
      \item Chronos-Bolt~\cite{ansari2024chronos} is proposed for pretrained probabilistic time
series models, which tokenize time series values using scaling and quantization into a fixed vocabulary. Existing transformer-based language model architectures are trained on these tokenized time series via the cross-entropy loss.

     \item  Time-MoE~\cite{DBLP:conf/iclr/ShiWNLYWJ25} introduces a MoE structure into the Transformer framework, enabling sparse activation where only a subset of experts is activated per input token.
      \item SEMPO~\cite{he2025sempo} is a light generalist time series forecasting model that was built in the frequency domain by modeling both high and low-energy frequency signal.
      \item DADA~\cite{DBLP:conf/iclr/ShentuL0SRPYG25} is the first generalist TSAD that is pre-trained on multiple domains datasets using adaptive bottlenecks and dual adversarial decoders.
 \end{itemize}

\begin{table*}[!t] \footnotesize
\setlength{\tabcolsep}{10pt}
 \centering
 \caption{Multi-metrics results in the three real-world datasets. The higher values for all metrics represent better performance. The best ones are in bold, and the second ones are underlined. }
 \label{tab:multi_metrics}
 \newcommand{\tabincell}[2]{\begin{tabular}{@{}#1@{}}#2\end{tabular}}
 % \scalebox{0.85}{
 \begin{tabular}{c|c|ccccccccc}
  \toprule
  \toprule
  \textbf{Dataset} & \textbf{Method} & \textbf{\textbf{Aff-P}} & \textbf{\textbf{Aff-R}} & \textbf{\textbf{Aff-F1}} & \textbf{AUC-R} & \textbf{AUC-P} & \textbf{R-A-R} & \textbf{R-A-P} & \textbf{V-R} & \textbf{V-P}\\
  \midrule 
  \multirow{7}{*}{\textbf{SMD}} & \textbf{Anomaly Transformer} & 54.08 & 97.07 & 66.42 & 36.74 & 03.87 & 51.22 & 07.09 & 51.17 & 07.96\\
   & \textbf{D3R} & 64.87 & \textbf{97.93} & 78.02 & 72.01 & 14.18 & 62.89 & 11.20 & 61.69 & 11.01 \\
  & \textbf{ModernTCN} & 74.07 & 94.79 & 83.16 & 70.21 & 14.95 & 77.54 & 16.28 & 77.07 & 15.96 \\
  & \textbf{GPT4TS} & 73.33 & 95.97 & 83.13 & \underline{74.15} & \underline{15.85} & 77.27 & \underline{17.68} & 76.79 & \underline{17.45} \\
  \cline{2-11}   
  & \textbf{TimesFM} & 61.44 & 88.47 & 72.52 & 62.80 & 07.37 & 69.72 & 12.08 & 74.58 & 12.69 \\
  & \textbf{Chronos-Bolt} & 52.69 & \underline{97.21} & 68.34 & 63.40 & 07.64 & 70.41 & 12.45 & 75.19 & 13.13 \\
  & \textbf{Time-MoE} & 60.85 & 88.84 & 72.23 & 61.94 & 07.25 & 68.93 & 11.74 & 73.82 & 12.37 \\  
  & \textbf{DADA} & \underline{76.50} & 94.53 & \underline{84.57} & 71.98 & 13.16 & \underline{78.14} & 17.39 & \underline{77.50} & 17.13\\
  & \textbf{TimeRadar} & \textbf{76.60} & 94.79 & \textbf{84.80} & \textbf{74.54} & \textbf{16.09} & \textbf{78.70} & \textbf{18.32} & \textbf{86.16} & \textbf{24.10} \\ 
  \midrule 
  \multirow{7}{*}{\textbf{MSL}} & \textbf{Anomaly Transformer} & 51.04 & 95.36 & 66.49 & 49.06 & 10.47 & 50.63 & 13.07 & 51.32 & 14.47 \\
   & \textbf{D3R} & 66.85 & 90.83 & 77.02 & 45.77 & 09.45 & 56.35 & 20.05 & 55.69 & 19.76 \\
  & \textbf{ModernTCN} & 65.94 & 93.00 & 77.17 & 63.04 & 15.20 & 77.88 & \underline{30.91} & 77.47 & 30.10 \\
  & \textbf{GPT4TS} & 64.86 & 95.43 & 77.23 & 58.85 & 13.83 & 77.65 & 28.05 & 76.97 & 27.69 \\
  \cline{2-11}
  & \textbf{TimesFM} & 59.68 & \underline{97.95} & 74.17 & 69.69 & 20.26 & 71.92 & 24.99 & 73.94 & 25.77 \\
  & \textbf{Chronos-Bolt} & 63.91 & 93.07 & 75.78 & 67.14 & 18.72 & 71.09 & 24.35 & 73.24 & 25.06 \\
  & \textbf{Time-MoE} & 49.59 & \textbf{98.86} & 66.05 & 61.04 & 16.79 & 69.48 & 23.14 & 71.65 & 23.62 \\
  & \textbf{DADA} & \textbf{68.70} & 90.56 & \underline{78.27} & \underline{75.15} & \textbf{26.42} & \underline{77.99} & \textbf{31.05} & \underline{77.48} & \underline{30.50} \\
  & \textbf{TimeRadar} & \underline{66.88} & 97.78 & \textbf{79.44} & \textbf{76.02} & \underline{24.38} & \textbf{79.66} & \textbf{31.05} & \textbf{81.63} & \textbf{35.95} \\ 
  \midrule 
  \multirow{7}{*}{\textbf{PSM}} & \textbf{Anomaly Transformer} & 54.26 & 82.18 & 65.37 & 49.99 & 28.05 & 52.10 & 33.53 & 52.86 & 33.05 \\
   & \textbf{D3R} & \underline{73.32} & 88.71 & 80.29 & 58.46 & 38.11 & 52.93 & 39.62 & 52.18 & 39.31 \\
  & \textbf{ModernTCN} & \textbf{73.47} & 86.83 & 79.59 & 58.91 & 38.26 & 65.50 & \underline{47.48} & 64.80 & 46.68 \\
  & \textbf{GPT4TS} & 73.61 & 91.13 & 81.44 & 56.26 & 35.82 & 65.16 & 46.54 & 64.66 & 45.99 \\
  \cline{2-11}
  & \textbf{TimesFM} & 53.37 & \textbf{99.99} & 69.58 & 51.04 & 29.30 & 53.38 & 34.90 & 54.56 & 34.78 \\
   & \textbf{Chronos-Bolt} & 53.15 & \underline{98.87} & 69.14 & 51.03 & 29.24 & 54.67 & 34.99 & 54.67 & 34.99 \\
   & \textbf{Time-MoE} & 53.91 & 95.89 & 69.02 & 54.68 & 32.31 & 60.75 & 38.52 & 62.28 & 38.50 \\
  & \textbf{DADA} & \textbf{73.47} & 91.71 & \underline{81.58} & \underline{61.08} & \underline{38.79} & \underline{67.08} & 47.40 & \underline{66.52} & \underline{46.89} \\
  & \textbf{TimeRadar} & 71.49 & 96.22 & \textbf{82.03} & \textbf{67.72} & \textbf{44.44} & \textbf{69.62} & \textbf{48.79} & \textbf{72.08} & \textbf{48.00} \\ 
  \bottomrule 
  \bottomrule
 \end{tabular}
 % }
\end{table*}

\begin{table*}[!t] \footnotesize
\setlength{\tabcolsep}{4.8pt}
 \centering
 \caption{Quantitative results for TimeRadar (\textit{Ours}) in five real-world datasets. The Aff-P, Aff-R and Aff-F1 represent the Affiliation Precision, Recall and F1 (as \%) respectively. The best ones are in bold, and the second ones are underlined.}
 \label{tab:quantitative}
 \newcommand{\tabincell}[2]{\begin{tabular}{@{}#1@{}}#2\end{tabular}}
 % \scalebox{0.7}{
 \begin{tabular}{c|ccc|ccc|ccc|ccc|ccc}
  \toprule
  \toprule
  \textbf{Dataset} & \multicolumn{3}{c}{\textbf{SMD}} & \multicolumn{3}{c}{\textbf{MSL}} & \multicolumn{3}{c}{\textbf{SMAP}} & \multicolumn{3}{c}{\textbf{SWaT}} & \multicolumn{3}{c}{\textbf{PSM}} \\
  \cmidrule(r){1-1} \cmidrule(r){2-4} \cmidrule(r){5-7} \cmidrule(r){8-10} \cmidrule(r){11-13} \cmidrule(r){14-16} 
  \textbf{Metric} & \textbf{Aff-P} & \textbf{Aff-R} & \textbf{Aff-F1} & \textbf{Aff-P} & \textbf{Aff-R} & \textbf{Aff-F1} & \textbf{Aff-P} & \textbf{Aff-R} & \textbf{Aff-F1} & \textbf{Aff-P} & \textbf{Aff-R} & \textbf{Aff-F1} & \textbf{Aff-P} & \textbf{Aff-R} & \textbf{Aff-F1} \\ 
  \midrule 
  \textbf{OCSVM} & 66.98 & 82.03 & 73.75 & 50.26 & 99.86 & 66.87 & 41.05 & 69.37 & 51.58 & 56.8 & 98.72 & 72.11 & 57.51 & 58.11 & 57.81 \\
  \textbf{PCA} & 64.92 & 86.06 & 74.01 & 52.69 & 98.33 & 68.61 & 50.62 & 98.48 & 66.87 & 62.32 & 82.96 & 71.18 & 77.44 & 63.68 & 69.89 \\
  \textbf{HBOS} & 60.34 & 64.11 & 62.17 & 59.25 & 83.32 & 69.25 & 41.54 & 66.17 & 51.04 & 54.49 & 91.35 & 68.26 & 78.45 & 29.82 & 43.21 \\
  \textbf{LOF} & 57.69 & 99.10 & 72.92 & 49.89 & 72.18 & 29.00 & 47.92 & 82.86 & 60.72 & 53.20 & 96.73 & 68.65 & 53.90 & 99.91 & 70.02 \\
  \textbf{IForest} & 71.94 & 94.27 & 81.61 & 53.87 & 94.58 & 68.65 & 41.12 & 68.91 & 51.51 & 53.03 & 99.95 & 69.30 & 69.66 & 88.79 & 78.07 \\
  \midrule
  \textbf{LODA} & 66.09 & 84.37 & 74.12 & 57.79 & 95.65 & 72.05 & 51.51 & 100.00 & 68.00 & 56.30 & 70.34 & 62.54 & 62.22 & 87.38 & 72.69 \\
  \textbf{AE} & 69.22 & 98.48 & 81.30 & 55.75 & 96.66 & 70.72 & 39.42 & 70.31 & 50.52 & 54.92 & 98.20 & 70.45 & 60.67 & 98.24 & 75.01 \\
  \textbf{DAGMM} & 63.57 & 70.83 & 67.00 & 54.07 & 92.11 & 68.14 & 50.75 & 96.38 & 66.49 & 59.42 & 92.36 & 72.32 & 68.22 & 70.50 & 69.34 \\
  \textbf{LSTM} & 60.12 & 84.77 & 70.35 & 58.82 & 14.68 & 23.49 & 55.25 & 27.70 & 36.90 & 49.99 & 82.11 & 62.15 & 57.06 & 95.92 & 71.55 \\
  \textbf{BeatGAN} & 74.11 & 81.64 & 77.69 & 55.74 & 98.94 & 71.30 & 54.04 & 98.30 & 69.74 & 61.89 & 83.46 & 71.08 & 58.81 & 99.08 & 73.81 \\
  \textbf{Omni} & 79.09 & 75.77 & 77.40 & 51.23 & 99.40 & 67.61 & 52.74 & 98.51 & 68.70 & 62.76 & 82.82 & 71.41 & 69.20 & 80.79 & 74.55 \\
  \textbf{CAE-Ensemble} & 73.05 & 83.61 & 77.97 & 54.99 & 93.93 & 69.37 & 62.32 & 64.72 & 63.50 & 62.10 & 82.90 & 71.01 & 73.17 & 73.66 & 73.42 \\
  \textbf{MEMTO} & 49.69 & 98.05 & 65.96 & 52.73 & 97.34 & 68.40 & 50.12 & 99.10 & 66.57 & 56.47 & 98.02 & 71.66 & 52.69 & 83.94 & 64.74 \\
  \textbf{Anomaly Transformer} & 54.08 & 97.07 & 66.42 & 51.04 & 95.36 & 66.49 & 56.91 & 96.69 & 71.65 & 53.63 & 98.27 & 69.39 & 54.26 & 82.18 & 65.37 \\
  \textbf{DCdetector} & 50.93 & 95.57 & 66.45 & 55.94 & 95.53 & 70.56 & 53.12 & 98.37 & 68.99 & 53.25 & 98.12 & 69.03 & 54.72 & 86.36 & 66.99 \\
  \textbf{SensitiveHUE} & 60.34 & 90.13 & 72.29 & 55.92 & 98.95 & 71.46 & 53.63 & 98.37 & 69.42 & 58.91 & 91.71 & 71.74 & 56.15 & 98.75 & 71.59 \\
  \textbf{D3R} & 64.87 & 97.93 & 78.02 & 66.85 & 90.83 & 77.02 & 61.76 & 92.55 & 74.09 & 60.14 & 97.57 & 74.39 & 73.32 & 88.71 & 80.29 \\
  \textbf{ModernTCN} & 74.07 & 94.79 & 83.16 & 65.94 & 93.00 & 77.17 & 69.50 & 65.45 & 67.41 & 59.14 & 89.22 & 71.13 & 73.47 & 86.83 & 79.59 \\
  \textbf{GPT4TS} & 73.33 & 95.97 & 83.13 & 64.86 & 95.43 & 77.23 & 63.52 & 90.56 & 74.67 & 56.84 & 91.46 & 70.11 & 73.61 & 91.13 & 81.44 \\
  \midrule
  \textbf{TimesFM} & 61.44 & 88.47 & 72.52 & 59.68 & 97.95 & 74.17 & 58.31 & 94.10 & 72.00 & 61.63 & 85.95 & 71.79 & 53.37 & 99.99 & 69.58 \\ 
  \textbf{Chronos-Bolt} & 52.69 & 97.21 & 68.34 & 63.91 & 93.07 & 75.78 & 62.24 & 90.97 & 73.91 & 56.19 & 96.53 & 70.98 & 53.15 & 98.87 & 69.14 \\
  \textbf{TimeMoE} & 60.85 & 88.84 & 72.23 & 49.59 & 98.86 & 66.05 & 56.86 & 94.04 & 70.87 & 52.81 & 99.58 & 69.02 & 53.91 & 95.89 & 69.02 \\
  \textbf{SEMPO} & 61.16 & 88.53 & 72.35 & 62.20 & 92.72 & 74.45 & 61.37 & 71.80 & 66.18 & 51.97 & 71.82 & 60.31 & 62.28 & 88.39 & 73.07 \\
  \textbf{DADA} & 76.50 & 94.53 & 84.57 & 68.70 & 90.56 & 78.27 & 66.53 & 83.69 & 74.13 & 59.22 & 97.54 & 73.70 & 73.47 & 91.71 & 81.58 \\
  \midrule
  \textbf{TimeRadar} & 76.60 & 94.79 & \textbf{84.80} & 66.88 & 97.78 & \textbf{79.44} & 62.90 & 94.14 & \textbf{75.41} & 62.23 & 97.40 & \textbf{76.73} & 71.49 & 96.22 & \textbf{82.03} \\
  \bottomrule 
  \bottomrule
 \end{tabular}
 % }
\end{table*}

  \begin{figure}
 \centering
  \includegraphics[width=0.48\textwidth]{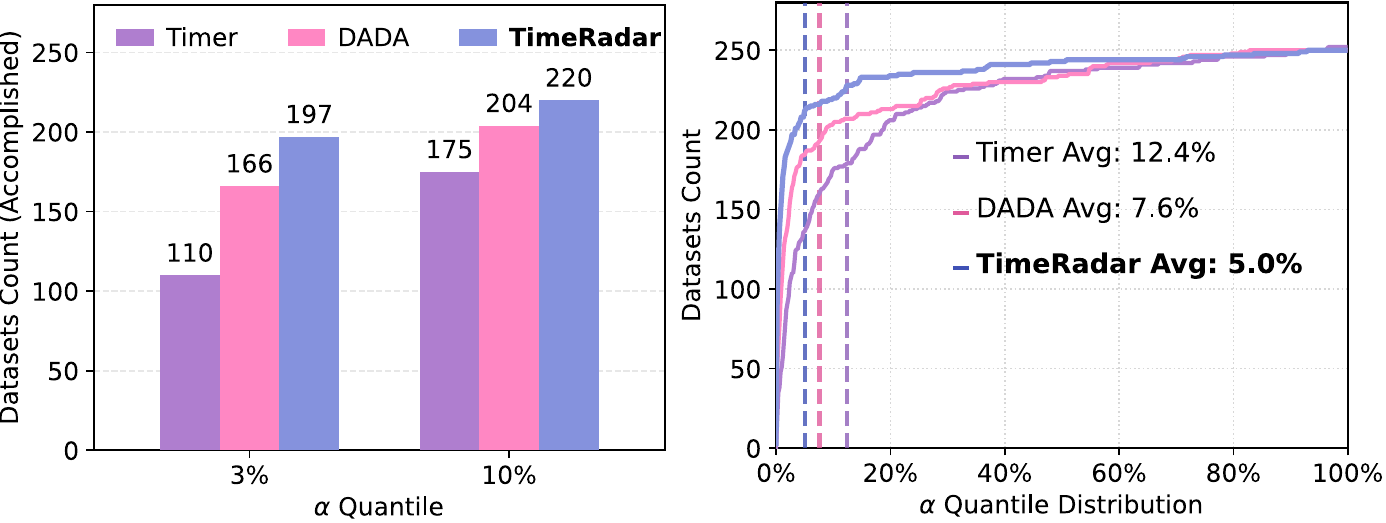}
 \caption{Results on UCR benchmark.}
 \label{fig7}
\end{figure}

\begin{figure*}[!t] 
\centering
\subfloat[Aff-F1 results w.r.t $T$]{
    \label{fig6a}
    \includegraphics[height=3.4cm]{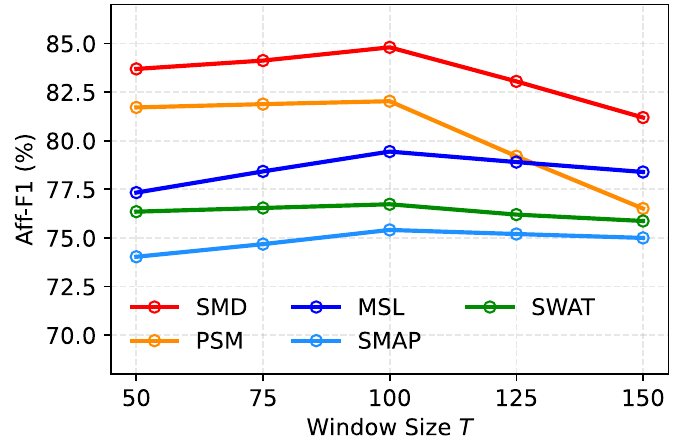}
    }
\hspace{0.2cm}
\subfloat[Aff-F1 results w.r.t $\gamma$]{
    \label{fig6b}
    \includegraphics[height=3.4cm]{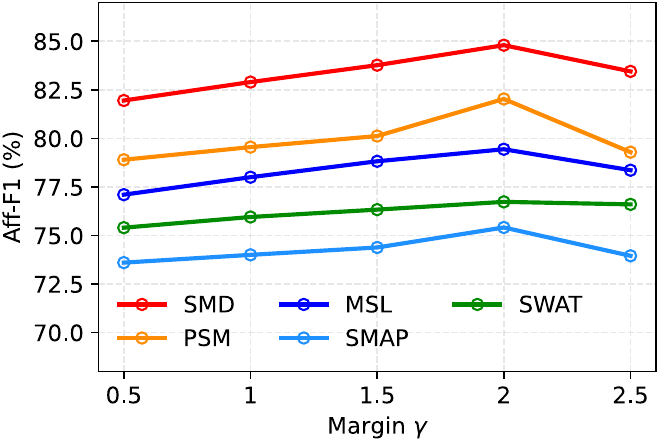}
    }  
\hspace{0.2cm}
\subfloat[Aff-F1 results w.r.t $\lambda$]{
    \label{fig6c}
    \includegraphics[height=3.4cm]{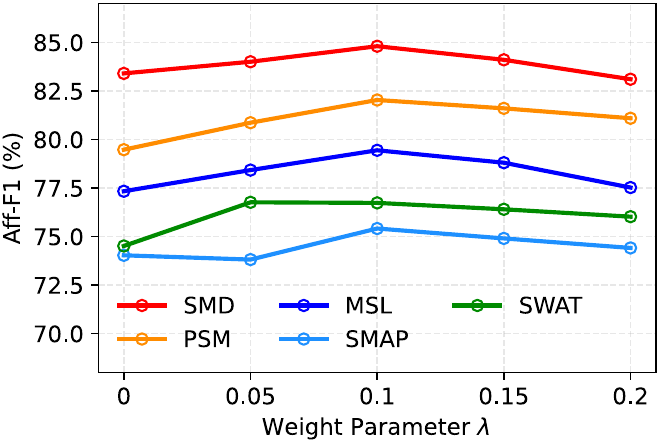}
    }
\centering
\caption{Sensitivity analysis w.r.t window size $T$, contextual deviation margin $\gamma$, and weight parameter $\lambda$.}
\label{fig:sensitive}
\end{figure*}

\section{Additional Experiments} \label{app:additional}
\subsection{Multi-Metric Results}\label{app:additional_metric}
Because different metrics provide complementary views of model performance and effectiveness, we report three widely used anomaly-detection metrics in the main text: Aff-F1, AUC-R, and AUC-P. To further strengthen the evaluation, we additionally include 6 metrics, with the description provided below and the results summarized in Table~\ref{tab:multi_metrics}. TimeRadar still achieves the best performance on all four additional metrics across the three datasets SMD, MSL and PSM, further demonstrating its effective anomaly detection capability.

\begin{itemize}
    \item Aff-P (Affinity-Precision) An affinity-based precision that measures the fraction of predicted abnormal sample affinities that are truly abnormal (a higher Aff-P indicates fewer false alarms when predicting abnormal affinities).
    \item Aff-R (Affinity-Recall) An affinity-based recall that measures the fraction of truly abnormal samples affinities that are successfully identified as abnormal (a higher Aff-R indicates better coverage of abnormal affinities).
    \item  R-A-R (Range-AUC-ROC): A range-aware ROC AUC that measures detection quality when anomaly samples are contiguous segments, giving credit for overlapping the true anomaly ranges (and being tolerant to small boundary misalignments). 
    \item R-A-P (Range-AUC-PR): A range-aware PR AUC that measures precision–recall performance under segment-based anomaly labeling, rewarding partial/complete coverage of anomaly ranges. 
    \item V-R (VUS-ROC): The Volume Under the ROC, which integrates (range-aware) ROC score over different window sizes, reflecting the robustness to timing misalignment. 
    \item V-P (VUS-PR): The Volume Under the PR, which integrates (range-aware) PR score over different window sizes, capturing robustness across detection tolerances. 

\end{itemize}

\subsection{Quantitative Results} \label{app:additional_quantitative}
Except for several recent strong competing baselines, we also include a set of earlier TSAD baselines for comprehensive comparison. These methods cover: (1) \textit{conventional methods}, including OCSVM~\cite{scholkopf1999support}, PCA~\cite{shyu2003novel}, HBOS \cite{goldstein2012histogram}, LOF \cite{breunig2000lof}, and IForest \cite{liu2008isolation}; and (2) \textit{deep learning methods}, including LODA~\cite{pevny2016loda}, AE \cite{sakurada2014anomaly},  DAGM \cite{zong2018deep}, LSTM~\cite{hundman2018detecting}, CAE-Ensemble \cite{campos2021unsupervised}, 
SensitiveHUE~\cite{feng2024sensitivehue}, BeatGAN~\cite{zhou2019beatgan}, and MEMTO~ \cite{song2023memto}. 
% are also included as the baselines for comparison in the evaluation. 
The experimental results are reported in Table \ref{tab:quantitative}.
TimeRadar achieves the best Aff-F1 across all five datasets, consistently outperforming both conventional methods and deep learning baselines. This indicates that TimeRadar attains a more favorable trade-off between precision and recall, making it better suited for real-world practical applications.

\subsection{Evaluation on the UCR Benchmark} \label{app:ucr}
In this section, following previous studies \cite{DBLP:conf/iclr/ShentuL0SRPYG25, wu2021current}, we evaluate TimeRadar on the UCR Anomaly Archive \cite{wu2021current}. The objective is to detect anomalous segments in the test split. For each timestep, anomaly scores are computed and ranked in descending order. If a labeled anomaly interval is detected at a time step with the $\alpha$ quantile, we regard the corresponding anomaly sample as successfully identified by the model. The evaluation results, including: (1) the number of successfully detected anomalies at quantile levels of 3\% and 10\%; and (2) the quantile distribution and the average quantile, are presented in Fig. \ref{fig7} \textbf{left} and \textbf{right}, respectively. We observe that under both the 3\% and 10\% quantiles, TimeRadar detects more anomalous samples than both DADA and Timer~\cite{DBLP:conf/icml/LiuZLH0L24}, indicating its stronger anomaly identification capability and more robust performance across varying quantile levels. As the quantile $\alpha$ increases, TimeRadar, DADA, and Timer identify (\ie, count) more anomaly samples, as expected. Nevertheless, TimeRadar consistently achieves the lowest average quantile compared with DADA and Timer, further indicating its superior effectiveness.

\begin{figure}
  \centering
  \includegraphics[width=0.47\textwidth]{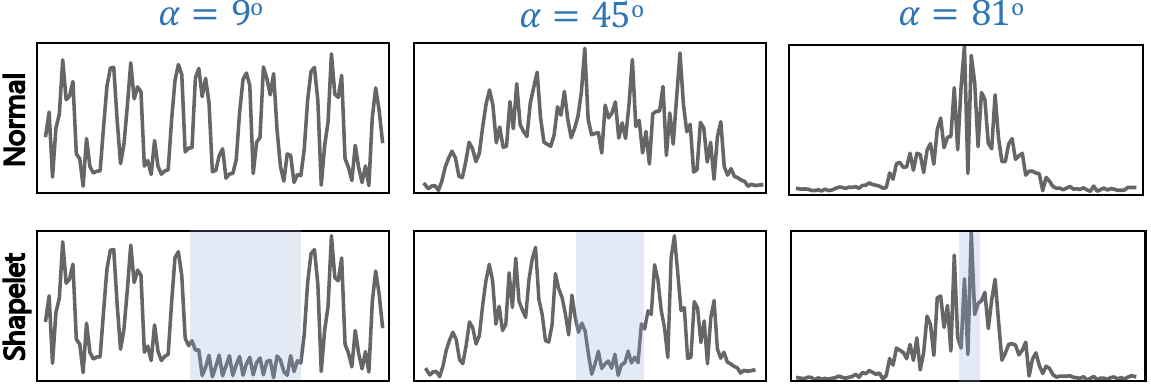}
  \caption{Visualization of the shapelet anomaly in Fig. \ref{fig:example} under the rotatable time–frequency domain with different rotation angles $\alpha$.}
  \label{fig:shapelet}
\end{figure}

\begin{figure}
\centering
\subfloat[Before training]{
    \label{fig3a}
    \includegraphics[height=2.9cm]{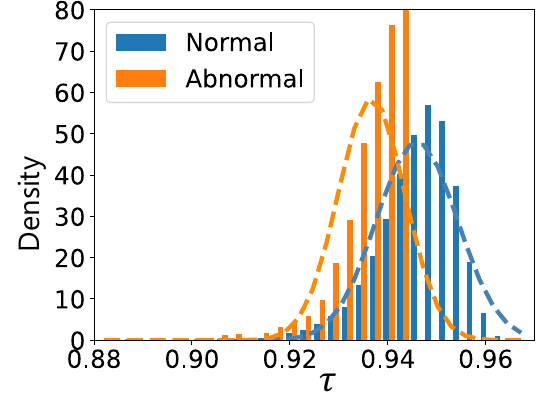}
    }
\subfloat[After training]{
    \label{fig3b}
    \includegraphics[height=2.9cm]{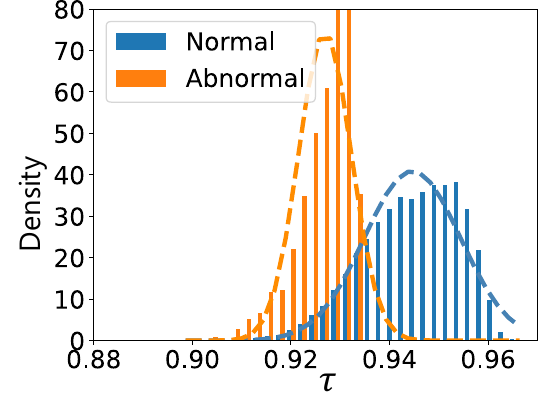}
    }
\centering
\caption{Contextual deviation differences: (a) before training and (b) after training. }
\label{fig:deviation}
\vspace{-1em}
\end{figure}

\section{Sensitive Analysis} \label{app:additional_sensitive}
% \begin{figure}[!t]
%  \centering
%  % Requires \usepackage{graphicx}
%  \includegraphics[width=0.48\textwidth]{figures/fig6.pdf}
%  \caption{Sensitivity analysis w.r.t  $T$ and $\gamma$.
% }
%  \label{fig:window_margin}
%  \vspace{-1em}
% \end{figure}

\noindent \textbf{The effectiveness of window size $T$.}
As shown in Fig.~\ref{fig:sensitive} (a), TimeRadar’s performance first improves slightly and then declines as the window size increases. The main reason is that with very short windows, the model sees limited temporal context, while when the window becomes too long, the abnormal signature becomes less salient in the learned representation. The best results are achieved when the window size is set to 100. Overall, TimeRadar remains stable across a wide range of window sizes, indicating low sensitivity to this hyperparameter.

\noindent \textbf{The effectiveness of contextual deviation margin $\gamma$.}
$\gamma$ in $L_{cdl}$ denotes the contextual difference we enforce between the normal patch and abnormal patch. As Fig.~\ref{fig:sensitive} (b) shows,  TimeRadar achieves strong performance on all five datasets across different values of $\gamma$, suggesting that enforcing contextual deviation is effective. The best overall results are obtained when $\gamma = 2$. However, $\gamma$ should not be set too large, since overly strict margins may cause hard anomalies with only subtle contextual deviations to be overlooked.
% as a slight performance drop is observed when $\gamma$ increases to 2.5.
% As illustrated in Fig.~\ref{fig:window_margin}, where  
Overall, TimeRadar remains stable under varying margin sizes.

\noindent \textbf{The effectiveness of weight parameter $\lambda$.}
As shown in Figs.\ref{fig:sensitive} (c), TimeRadar generally stable as the weight $\lambda$ varies. The best overall results are obtained when $\lambda = 0.1$. As $\lambda$ increases, the performance exhibits a slight decline. This is likely excessive emphasis on contextual deviation learning can overly compress the representation space, which in turn reduces the discriminative power of the time–frequency representations.

\section{More Visualization of Shapelet Anomalies in Rotatable Time–Frequency Domain} \label{app:anomaly_visualization}
We provide the visualization of shapelet anomalies with varying rotation angles, as illustrated in Fig.~\ref{fig:shapelet}. The discrepancy between anomalous shapelets and normal patterns changes significantly with different rotation angles. When the rotation angle approaches $81^{\rm o}$, which can be regarded as a near-frequency-domain setting, the anomalous patterns become less salient. As the angle decreases, the distinction becomes increasingly pronounced, making the anomalies easier to identify.

\section{Visualization of Contextual Deviation Difference} \label{app:deviation}

We visualize the contextual deviation difference before applying CDL and after CDL training to assess its effectiveness, as shown in Fig.\ref{fig:deviation}. Applying CDL amplifies the contextual deviation contrast, suggesting that CDL enhances contextual deviation contrast and yields a more distinct boundary between normal and abnormal time series.

\end{document}